\documentclass{article} 
\usepackage{iclr2026_conference,times}

\usepackage{amsmath,amsfonts,bm}

\newcommand{\ours}{\textsf{HI-RL}\;}

\def\eqref#1{equation~\ref{#1}}

\def\1{\bm{1}}

\DeclareMathAlphabet{\mathsfit}{\encodingdefault}{\sfdefault}{m}{sl}
\SetMathAlphabet{\mathsfit}{bold}{\encodingdefault}{\sfdefault}{bx}{n}

\usepackage{hyperref}
\usepackage{url}

\usepackage{amsmath}
\usepackage{amssymb}
\usepackage{mathtools}
\usepackage{amsthm}
\usepackage{wrapfig}

\usepackage{algorithm}
\usepackage{algpseudocode}
\usepackage{booktabs}

\theoremstyle{plain}
\newtheorem{theorem}{Theorem}[section]

\newtheorem{lemma}[theorem]{Lemma}

\theoremstyle{definition}
\newtheorem{definition}[theorem]{Definition}

\theoremstyle{remark}

\definecolor{laixi}{RGB}{255,125,0} 

\newcommand{\cA}{\mathcal{A}}

\newcommand{\mymid}{\,|\,} 

\usepackage{scalerel,stackengine}
\stackMath
\newcommand\reallywidehat[1]{%
\savestack{\tmpbox}{\stretchto{%
  \scaleto{%
    \scalerel*[\widthof{\ensuremath{#1}}]{\kern-.6pt\bigwedge\kern-.6pt}%
    {\rule[-\textheight/2]{1ex}{\textheight}}
  }{\textheight}%
}{0.5ex}}%
\stackon[1pt]{#1}{\tmpbox}%
}
\newcommand\reallywidecheck[1]{%
\savestack{\tmpbox}{\stretchto{%
  \scaleto{
    \scalerel*[\widthof{\ensuremath{#1}}]{\kern-.6pt\bigwedge\kern-.6pt}%
    {\rule[-\textheight/2]{1ex}{\textheight}}
  }{\textheight}%
}{0.5ex}}%
\stackon[1pt]{#1}{\scalebox{-1}{\tmpbox}}%
}

\iclrfinalcopy
\title{Conceptual Belief-Informed Reinforcement Learning}

\author{Xingrui Gu \\
University of California, Berkeley\\
\texttt{xingrui\_gu@berkeley.edu} \\
\And
Chuyi Jiang \\
Columbia University \\
\texttt{cj2792@columbia.edu} \\
\AND
Laixi Shi \\
Johns Hopkins University\\
\texttt{laixis@jhu.edu}
}

\begin{document}

\maketitle

\begin{abstract}

Reinforcement learning (RL) has achieved significant success but is hindered by inefficiency and instability, relying on large amounts of trial-and-error data and failing to efficiently use past experiences to guide decisions. However, humans achieve remarkably efficient learning from experience, attributed to abstracting concepts and updating associated probabilistic beliefs by integrating both uncertainty and prior knowledge, as observed by cognitive science.
Inspired by this, we introduce Conceptual Belief-Informed Reinforcement Learning to emulate human intelligence (HI-RL), an efficient experience utilization paradigm that can be directly integrated into existing RL frameworks. HI-RL forms concepts by extracting high-level categories of critical environmental information and then constructs adaptive concept-associated probabilistic beliefs as experience priors to guide value or policy updates. We evaluate HI-RL by integrating it into various existing value- and policy-based algorithms (DQN, PPO, SAC, and TD3) and demonstrate consistent improvements in sample efficiency and performance across both discrete and continuous control benchmarks.


\end{abstract}

\section{Introduction}



Reinforcement Learning (RL) has achieved remarkable success in various exciting areas, including aligning and enabling efficient inference of large language models \cite{ouyang2022training,hao2025rl}, game playing \citep{mnih2015human}, robotics \citep{singh2022reinforcement}, autonomous driving \citep{kiran2021deep}, and etc. Despite these achievements, RL remains fundamentally limited by its significant sample inefficiency compared to human learning \citep{chiu2023flexible,ye2021mastering, joshi2025shire}, typically relying on vast amounts of trial-and-error interactions and often struggling to generalize to unseen or sparsely observed space (states) \citep{mnih2015human, lake2017building}. In contrast, humans can quickly learn and adapt to new spaces using only a handful of experiences, highlighting a substantial gap in data efficiency between RL and human cognition \citep{tenenbaum2006theory, lake2015human, tenenbaum2011grow, griffiths2010probabilistic}.


The gap in learning efficiency motivates the “Era of Experience” \citep{silver2025welcome}, which emphasizes leveraging past interactions to accelerate learning and foster new concepts and behaviors, rather than passively processing vast amounts of data. Cognitive science highlights two mechanisms for leveraging experience that are essential to human learning efficiency \citep{tenenbaum2011grow}: \emph{conceptual abstraction} and \emph{probabilistic priors}. Conceptual abstraction distills reusable structures such as prototypes, taxonomies, causal schemas --- that enable compositional reasoning, generalization, and knowledge transfer \citep{tenenbaum2011grow, lake2015human, rosch1978principles, kemp2008discovery}. In parallel, behavioral studies show that humans aggregate past experiences into adaptive probabilistic priors \citep{griffiths2005structure, peterson1967man}, integrating them with future uncertainty to guide predictions and decisions \citep{griffiths2005structure, tenenbaum2006theory}. 

\begin{figure}[t]
    \centering
    \includegraphics[width=1.0\textwidth]{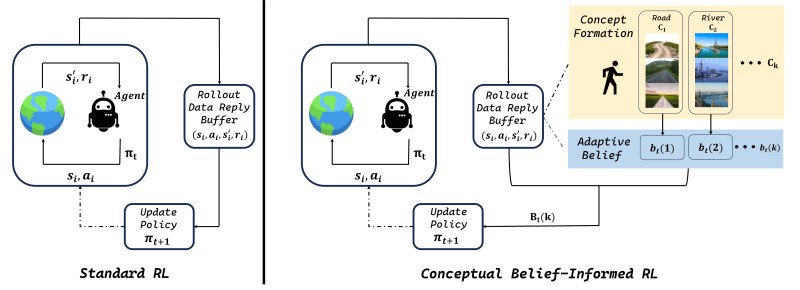}
    \caption{Standard RL (left) replays raw transitions, while HI-RL (right) organizes them into \textbf{conceptual categories} with \textbf{adaptive beliefs}, enabling abstraction and belief-guided learning.}
    \label{fig:introgai}
\end{figure}

Various RL studies leveraged either conceptual abstraction or probabilistic priors from experience independently. However, a systematic approach to combining both experience utilization mechanisms - experience-based priors grounded in extracted conceptual formations, the very mechanism underlying humans' efficient generalization—remains underexplored \citep{gerstenberg2017intuitive}. Specifically, abstraction in RL has focused on representation learning approaches such as contrastive learning and bisimulation metrics to compress or align observations into compact latent spaces to improve downstream task efficiency \citep{patil2024contrastive, ferns2004metrics, castro2020scalable, peng2023conceptual}. However, these methods typically do not further exploit the abstracted latent space to aggregate past experience, limiting its utility for improving RL learning efficiency. In parallel, Bayesian approaches, such as Thompson sampling, Bayesian model-based, and model-free algorithms \cite{dearden1998bayesian} are widely used to address uncertainty and the exploration-exploitation tradeoff in RL \citep{dearden1998bayesian, ghavamzadeh2015bayesian, ross2008model, thompson1933likelihood, dearden1998bayesian}. However, these methods are rarely integrated with conceptual abstraction. 

To efficiently leverage experience, we introduce Conceptual Belief-Informed RL, named \ours (Human Intelligence-RL), a framework that combines conceptual abstraction and concept-based probabilistic prior, illustrated in Figure~\ref{fig:introgai}. \ours provides an algorithm-agnostic interface that integrates seamlessly with existing RL frameworks, accelerating learning by leveraging experience efficiently. It extracts concepts from large state spaces and reformulates experience into priors grounded in these abstractions, mimicking human-like conceptualization for learning efficiency. Our main contributions are summarized as below:

\begin{itemize}  
    \item We present \ours, an experience-utilization framework that efficiently leverages past experiences to emulate human-like learning efficiency. \ours reformulates the set of past experiences into probabilistic belief priors grounded in conceptual abstractions. These concept-based priors are adaptively updated over time and incorporated as auxiliary knowledge into RL value or policy updates.

    \item \ours is algorithm-agnostic and functions as a flexible module that can be seamlessly integrated into existing RL frameworks. To demonstrate its versatility, we integrate \ours into several popular RL algorithms (Q-learning, PPO, SAC, and TD3) and evaluate performance across both discrete and continuous tasks, achieving consistent improvements in learning efficiency and overall performance.

\end{itemize}

\section{Related Works}



\subsection{Cognitive Science for Conceptual Learning}

Humans achieve remarkable learning efficiency by generalizing from limited experience through Bayesian inference, integrating prior knowledge with new evidence under uncertainty \citep{tenenbaum2001generalization, griffiths2005structure, tenenbaum2006theory}. This supports conceptual abstraction—extracting high-level structure from sparse data—and enables causal reasoning and cross-domain transfer \citep{tenenbaum2011grow, kemp2008discovery}. Recent work formalizes how learners reorganize internal knowledge via probabilistic reasoning \citep{lake2015human, lake2017building}, motivating the integration of such principles into machine learning for scalability, adaptability, and sample efficiency \citep{ma2022learning}. Studies further show that uncovering latent causal structures enhances interpretability and abstraction, even in complex domains such as joint behavior analysis \citep{gu2025causkelnet, gu2024advancing}. Yet, despite these advances, reinforcement learning remains dominated by replay, metric-based similarity, or policy integration, with little use of structured conceptual abstraction from cognitive science.



\subsection{Experience-Informed Reinforcement Learning}
Experience has long been exploited to improve efficiency in RL. Habit-based RL models long-term regularities as habitual priors that accelerate action selection but lack flexibility for abstraction and transfer \citep{daw2005uncertainty, collinsbeyond, keramati2011speed}. Replay-based techniques such as PER \citep{schaul2015prioritized}, HER and its prioritized variants \citep{andrychowicz2017hindsight, sun2025hierarchical, kim2025extended} enhance sample efficiency by weighting or relabeling transitions, while refinements like FoDA \citep{chen2024foresight} and EDER \citep{zhao2024efficient} adapt distributions or promote diversity to improve generalization. Beyond replay, episodic memory models (NEC) \citep{pritzel2017neural} enable rapid value retrieval, and hybrid gradients (Q-Prop, IPG) \citep{gu2016q, gu2017interpolated} fuse on- and off-policy signals for variance reduction. Collectively, these methods leverage past interactions via sampling or memory mechanisms, yet remain confined to buffer-level operations and lack pathways for higher-order conceptual abstraction and belief-structured generalization.

\subsection{Abstraction in Reinforcement Learning}

State abstraction has long been studied as a means to compress state spaces and enable generalization in RL \citep{bertsekas1988adaptive, givan2003equivalence, ravindran2004algebraic, ravindran2003smdp, li2006towards, kulkarni2016hierarchical}. Classical bisimulation and Kantorovich metrics provide strong theoretical guarantees but are computationally expensive and highly sensitive to perturbations \citep{ferns2004metrics, ferns2011bisimulation}. Task-specific metrics improve offline evaluation \citep{pavse2023state} but lack adaptability, while scalable relaxations \citep{castro2020scalable} trade rigor for tractability. Trajectory-chain and pseudometric methods \citep{girgin2007state, dadashi2021offline} offer finer granularity but incur high storage or auxiliary costs. More recent work, such as \citep{patil2024contrastive}, leverages contrastive objectives and modern Hopfield networks to compress large state spaces into abstract nodes, thereby facilitating downstream RL. These approaches primarily focus on constructing a new, compressed state space or representation for downstream algorithms. In contrast, our framework preserves the original state and exploration space while introducing an abstraction-based belief layer on top. We focus on utilizing conceptual abstraction as a basis to update its probabilistic priors, efficiently aggregating and using past experience to improve toward human-like efficient learning.

\section{Problem Formulation}
\textbf{Markov Decision Process (MDP) } 
Considering reinforcement learing problems formalized as MDP \citep{bellman1957markovian,sutton2018reinforcement} $\mathcal{M} = (\mathcal{S}, \mathcal{A}, \mathcal{T}, r, \mu_0, \gamma, T)$. Here $T$ is the horizon length.  $\mathcal{S}$ denotes states space ($s \in S$) and $\mathcal{A}$ denotes the action spaces ($a\in A$). $\mathcal{T}(s_{t+1}\mid s_t,a_t)$ represents the transition dynamics, specifying the probability distribution over the next state $s_{t+1}$ conditioned on the current state $s_t \in \mathcal{S}$ and action $a_t \in \mathcal{A}$ at $t_{th}$ time step $(1 \leq t \leq T)$. $r(s,a)$ represents the reward function given the state s and action a.
The initial state follows $\mu_0$, $\gamma \in (0,1)$ is the discount factor. 
The goal of this MDP problem is to identify an optimal policy $\pi$ that achieves the maximum expected discounted return:
\begin{align}
    \max_{\pi}\; \mathbb{E}_{\pi,\mathcal{T},\mu_0}\Big[\sum_{t=0}^{T}\gamma^t r(s_t,a_t)\Big].
\end{align}
Formally, given the horizon length $T$ and transition dynamics $\mathcal{T}$, the long-term return from time step $t = 0$ to $t = T$ associated with the optimal policy $\pi$ is quantified through the Q-function and the Value-function \citep{watkins1992q} by expected cumulative rewards from initial state $\mu_0$, defined as: 
\begin{align}
    Q^\pi(s,a) = \mathbb{E}_{\pi,\mathcal{T},\mu_0}\Big[\sum_{t=0}^T \gamma^t r(s_t,a_t)\;\big|\; s_0=s,a_0=a\Big], 
    V^\pi(s) = \mathbb{E}_{\pi,\mathcal{T},\mu_0}\Big[\sum_{t=0}^T \gamma^t r(s_t,a_t)\;\big|\; s_0=s\Big]
\end{align}
\section{Conceptual Belief-Informed Reinforcement Learning}


In this section, we present Conceptual Belief-Informed Reinforcement Learning (\ours), enhancing experience to emulate human intelligence learning efficiency. \ours consists of two core modules: (i) {\em Concept Formation}, which clusters state–based experiences into semantically coherent categories, and (ii) a {\em belief} representing probabilistic prior grounded on different concepts, defining probabilistic action experience prior over these categories. By coupling conceptual abstraction with belief-guided reasoning, \ours provides a structured and uncertainty-aware foundation for policy learning, supporting stable updates, efficient generalization, and reuse of past experiences.




\subsection{Concept Formation}

The foundation for abstracting concepts can vary, as long as it represents the current situation and critical information about the environment and the agent. In this work, \ours focuses on state spaces, as states encapsulate essential information for decision-making and directly influence the agent's behavior and learning process, enabling pattern recognition and generalization. Specifically, we partition the states into disjoint subsets, with each subset representing a distinct concept formed by grouping states with shared characteristics and properties, thereby facilitating effective knowledge transfer within each concept. In the following, we mathematically define a conceptual abstraction as:

\begin{definition}[Concept Formation in State Space]
A concept formation in the state space is defined as a collection of subsets 
$C_K = \{C_1, \dots, C_K\}$ that satisfy $\mathcal{S} = \bigcup_{k=1}^K C_k$, meaning the subsets are disjoint and collectively cover the entire state space. 
Here, $K$ denotes a finite, prescribed number of concept categories.
\label{ccf}
\end{definition}

In practice, conceptual abstractions can be obtained with various clustering methods. In this work, we adopt K-means \citep{lloyd1982least} for its simplicity and scalability, though alternatives (e.g., spectral or hierarchical clustering) are equally applicable.

\subsection{Conceptual Adaptive Belief for RL}


With abstract concepts in mind, where each concept groups states that share similar features and actions, we aggregate observed information within a concept into a unified container, a {\em concept-based belief}. Philosophically, a {\em belief} is an internal representation of how an agent interprets and anticipates the world, serving as a guide for inference and decision-making under uncertainty rather than as absolute truth \citep{dennett1988precis}. In this work, each concept $C_k$ is paired with a time-adaptive belief $b_t(\cdot \mid k) \in \Delta(\mathcal{A})$, derived from the accumulation of past decisions and outcomes within that concept. Formally, for a conceptual abstraction $C_K = \{C_1,\dots,C_K\}$, we define the mapping $b_t:[K]\to\Delta(\mathcal{A})$, where $b_t(\cdot\mid k)$ encodes the integrated action preferences of all states belonging to $C_k$.


We leverage the aggregated experience within each concept to accelerate learning by using concept-based beliefs as priors in RL updates. These beliefs can be seamlessly integrated into any existing RL algorithm. At each timestep $t$, we combine two signals: (i) instant feedback $\mathcal{Z}_t:\mathcal{S}\to\mathcal{A}$, defined by the base algorithm (e.g., Q-values in DQN, Gaussian policy in SAC, clipped surrogate in PPO, or deterministic actor in TD3), and (ii) the prior $b_t$, aggregated from past experience within the corresponding concept. For a given state $s \in \mathcal{S}$, we first identify its concept index $c(s)$ such that $s \in C_{c(s)}$, and then fuse the signals as:
\begin{equation}
   B_t(\cdot \mid s) = (1 - \beta_t) \mathcal{Z}_t(\cdot \mid s) + \beta_t b_t (\cdot \mymid c(s) ),
   \label{ceof}
\end{equation}
where $\beta_t \in [0,1]$ is an adaptive parameter monotonic in $t$, satisfying $\lim_{t \to \infty} \beta_t = \beta^*$ with $\beta^* \in [0,1]$ a constant denoting the limiting weight on conceptual priors. In this formulation, $b_t(\cdot \mid c(s))$ is the empirical concept-based prior aggregated from experience, while $B_t(\cdot \mid s)$ is the fused distribution actually used for decision-making by combining $b_t$ with the instant feedback $\mathcal{Z}_t$.

This formulation ensures that the decision-making solutions for every state $s_t$ are influenced by both immediate feedback from the environment and the prior experience derived from the conceptual abstraction $C_{c(s)}$ to which it belongs.


\begin{algorithm}[H]
\caption{Conceptual Belief-Informed RL (HI-RL)}
\label{alg:cbdrl}
\begin{algorithmic}[1]
\State Initialize concept priors $b(\cdot\mid c(s))$
\For{$t=1,2,\dots$}
    \State Observe $s_t$; form $\mathcal{Z}_t(\cdot\mid s_t)$; fuse $B_t(\cdot\mid s_t)=(1-\beta_t)\mathcal{Z}_t(\cdot\mid s_t)+\beta_t b_t(\cdot\mid c(s_t))$
    \State Sample $a_t\!\sim\!B_t$; step env $(r_t,s_{t+1})$
    \State Update policy with $B_t$ and prior experience $b_t$
\EndFor
\end{algorithmic}
\end{algorithm}

\section{Algorithm Implementation}
We apply \ours framework into multiple RL paradigms by developing HI-Q, HI-PPO and HI-SAC (For HI-TD3, see Appendix~\ref{al:td3_implement}).

\subsection{Conceptual Belief-Informed Q-learning (HI-Q)}
The classical Deep Q-learning (DQN) algorithm \citep{mnih2015human} relies on updating the Q-function network using the greedy Bellman operator. Namely, in any iteration $t$, with the sampled batch $D_t$ and any sample $(s_i, a_i, r_i, s_i') \in D_t$ within it, the learning target of the Q-network $Q_{\theta_{t+1}}(s_i, a_i)$ would be
\begin{equation}\label{eq:q-learning}
    r_i + \gamma \max_{a\in\mathcal{A}} Q_{\theta_t}(s_i',a_i),
\end{equation}
where $\theta_t$ represent the Q-network parameter at time $t$.
With DQN in mind, we propose HI-Q to replace the learning target to a new one combining both the current Q-network information $Q_{\theta_t}$ and the conceptual abstraction experience prior $b_t$.

Specifically, we first introduce the construction of the concept-based belief prior $b_t(\cdot \mymid k)$ at each time step $t$. Here, $b_t(\cdot \mymid k)$ will be defined as the action visiting frequency summarzied over all state within the concept set $C_k$. For the discrete finite action space $\cA$, we denote the number of visiting time over each state-action pair at time step t as $N_t(s,a)$. Then the experience prior $b_t$ of any $k$-th concept will be constructed as
\begin{equation}\label{eq:discrete-belief}
\forall (a,k) \in \cA \times [K]: \quad b_{t}(a \mid k) = \frac{ \sum_{s \in C_k} N_t(s,a) }{\sum_{a'\in\cA} \sum_{s \in C_k} N_t(s,a') }.
\end{equation}
The update of $b_t$ is typically computational easily, since upon executing an sample tuple $(s_i, a_i, r_i, s_i')$, only the $(s_i,a_i)$-associated concept $b_t(a_i \mymid c(s_i))$ will be updated.

Therefore, the combined information for any sample tuple $(s_i, a_i, r_i, s_i')$ associated with state $s_i'$ at time $t$ is defined as
\begin{align}
    \label{eq:Q-beta}
     B_t(\cdot \mid s_i') = (1 - \beta_t) q_t(\cdot \mid s_i') + \beta_t b_t (\cdot  \mymid c(s_i') ),
\end{align}
where $\beta_t$ is a dynamic coefficient and 
$q_t(\cdot \mid s'_i)$ denotes the \emph{task-driven action-preference distribution}, 
typically instantiated as a smoothing distribution over $Q$-values (e.g., softmax with temperature $\tau_t$ or clipped-max with exploration mass $\delta_t$) that gradually concentrates on the greedy action as $t$ increases \citep{barber2023smoothed}, computed via a softmax over the top-$k$ Q-values of state $s'_i$, effectively assigning higher probabilities to the most promising actions:
\begin{equation}
q_t(a \mid s'_i) = 
\frac{\exp(Q(s'_i, a)/\tau)}{\sum_{a' \in \text{top-}k(s'_i)} \exp(Q(s'_i, a')/\tau)}, \quad a \in \text{top-}k(s'_i)
\label{eq:Q-q_t}
\end{equation}
where $\tau$ is softmax temperature constant. With the constructed concept-based belief based template $ B_t$ in hand, we replace the (greedy) maximum operator in Eq.~\ref{eq:q-learning} of classical Q-learning to a smoothed surrogate one combining both the smoothed-greedy operator of the current Q-function and the conceptual-based belief. Therefore, the new target in HI-Q for the Q-network to learn is defined as
\begin{align}
    r_i + \gamma \sum_{a \in \mathcal{A}} B_t(a \mid s_i') Q_t(s_i', a).
    \label{eq:q-target}
\end{align}
The entire algorithm is specified in Appendix~\ref{al:Q}. Our conceptual-abstraction belief enables HI-Q to leverage both immediate task feedback and accumulated conceptual structures, facilitating faster learning by borrowing experience from other similar concepts.

\subsection{Conceptual Belief-Informed Proximal Policy Optimization (HI-PPO)}
The standard PPO \citep{schulman2017proximal} is a policy gradient algorithm which updates the policy by performing stochastic gradient ascent on a surrogate objective function. For any time step $t$, let $D_t$ be a sampled batch and $(s_i, a_i, r_i, s_i') \in D_t$ any individual sample within it. 
The objective is to update the policy $\pi_{\theta}(a \mid s)$ via the following loss function:
\begin{equation}
\mathcal{L}_{\text{PPO}} = \mathbb{E}_{(s_i,a_i)\sim\pi_{\theta_{\text{old}}}} \Big[ \min \Big( \tfrac{\pi_{\theta}(a_i \mid s_i)}{\pi_{\theta_{\text{old}}}(a_i \mid s_i)} A_t,\; \text{clip}\big(\tfrac{\pi_{\theta}(a_i \mid s_i)}{\pi_{\theta_{\text{old}}}(a_i \mid s_i)}, 1 - \epsilon, 1 + \epsilon \big) A_t \Big) \Big].
\label{eq:cbdppo_clip}
\end{equation}
where $\theta$ denotes the policy parameters, $A_t$ is the advantage estimate at time step $t$, and $\epsilon$ controls the trust region. While PPO updates the policy via an advantage-weighted likelihood ratio within this trust region, it depends only on immediate feedback, limiting its ability to exploit structural regularities. To overcome this, HI-PPO integrates the current policy $\pi_{\theta}(a \mid s)$ with the conceptual abstraction prior $b_t$.


In this paper, we focus on applying PPO in discrete action-space environments, with modifications analogous to those in HI-Q. At each time step $t$, we compute a concept-based belief prior $b_t(\cdot \mid k)$, defined as the action visitation frequency aggregated over all states in concept set $C_k$. Its computation and update follow Eq.~\ref{eq:discrete-belief}, and it is combined with the policy $\pi_{\theta}(\cdot \mid s_i)$ for state $s_i \in D_t$ at time $t$ as:
\begin{equation}
B_t(\cdot \mid s_i) = (1 - \beta_t) \pi_{\theta}( \cdot \mid s_i) + \beta_t b_t(\cdot \mid c_k(s_i)),
\label{eq:cbdppo_beta}
\end{equation}
where the scheduling parameter \(\beta_t \in [0, 1]\) controls the influence of concept priors and increases gradually throughout training. The clipped surrogate objective of HI-PPO is: 
\begin{equation}
\mathcal{L}_{\text{HI-PPO}} = \mathbb{E}_{(s_i,a_i)\sim\pi_{\theta_{\text{old}}}} \Big[ \min \Big( \tfrac{B_t(a_i \mid s_i)}{\pi_{\theta_{\text{old}}}(a_i \mid s_i)} A_t,\; \text{clip}\big(\tfrac{B_t(a_i \mid s_i)}{\pi_{\theta_{\text{old}}}(a_i \mid s_i)}, 1 - \epsilon, 1 + \epsilon \big) A_t \Big) \Big].
\label{eq:cbdppo_loss}
\end{equation}
The critic and entropy terms follow the original PPO formulation; gradients are propagated through \(B_t\), allowing concept priors to steer policy updates while the clip operator guarantees trust-region stability. More implementation details and pseudocode are provided in Appendix \ref{al:ppo}.

\subsection{Conceptual Belief-Informed Soft Actor-Critic (HI-SAC)}
Traditional Soft Actor-Critic (SAC) is a maximum entropy reinforcement learning algorithm that integrates both an actor and a critic network \citep{haarnoja2018soft}. Given a sampled batch $D_t$ at time step $t$, containing tuples $(s_i, a_i, r_i, s_i')$, the updates of the actor and critic networks parameters $\theta$ , $\phi$ from $\pi_{\theta}$ and $Q_{\phi}$ respectively are defined as follows:
\begin{equation}
\label{eq:sacupdate}
\begin{aligned}
\mathcal{L}_{\text{critic}}(\phi_i) &= 
\mathbb{E}\Big[ \big(Q_{\phi_i}(s_i,a_i) - y_t \big)^2 \Big], \quad i=1,2, \\
\text{where}& \quad
y_t = r_i + \gamma \, \mathbb{E}_{a_i' \sim \pi_\theta}
\big[ Q_{\min}(s_i', a_i') - \alpha \log \pi_\theta(a_i' \mid s_i') \big], \\
\mathcal{L}_{\text{actor}}(\theta) &= 
\mathbb{E}_{s_i \sim \mathcal{D}_t,\, a_i \sim \pi_\theta}
\big[ \alpha \log \pi_\theta(a_i \mid s_i) -  \min\{Q_{\phi_1}(s_i,a_i), \, Q_{\phi_2}(s_i,a_i)\} \big].
\end{aligned}
\end{equation}
where $\alpha$ is entropy temperature coefficient, $y_t$ is the TD target, computed by the next state $s_i'$ at time step $t+1$ and the corresponding action $a_i'$ sampled from the policy $\pi_{\theta}$. In SAC, the Q value is computed as the minimum of the estimates from two critic networks $Q_{\phi_i}$ and the actor network produces a Gaussian policy in this paper: 
\begin{equation}
\pi_{\theta}(\cdot \mid s) = \mathcal{N}(\mu_{\pi_\theta}(s), \sigma^2_{\pi_\theta}(s)),
\end{equation}
where $\mu_{\pi_\theta}(s)$ and $\sigma^2_{\pi_\theta}(s)$ denote the mean and variance predicted by the policy network for state $s$. To support concept-informed decision-making in continuous action spaces, we propose HIS-AC to integrate current actor network $\pi_{\theta_t}$ and the conceptual experience prior $b_t$. Unlike HI-Q and HI-PPO, the concept-based belief prior $b_t(k)$ constructed at each time step t is defined over the actor network parameters corresponding to the states s within the concept set $C_k$:
\begin{equation}\label{eq:continous-belief}
\forall (\mu,\sigma^2,k) \in  \{\mu,\sigma^2\} \times [K]: \quad b_{t}(k) = \{\mu_{\pi_{\theta}}(s),\sigma^2_{\pi_{\theta}}(s)\}, \quad s \in C_k
\end{equation}
During training, we update $b_t$ using a Bayesian posterior update. Let $\mu_c$ and $\sigma_c^2$ be the parameters of the current policy $\pi_{\theta_t}(s)$ and $\mu_e$ and $\sigma_e^2$ be the experience stored in $b_{t-1}(k)$:
\begin{equation}
 b_t(k)  = \{\frac{\sigma_c^2 \, \mu_e + \sigma_e^2 \, \mu_c}{\sigma_c^2 + \sigma_e^2},  \frac{1}{\frac{1}{\sigma_c^2} +  \frac{1}{\sigma_e^2}}\},\quad (\mu_c, \sigma_c^2) \sim \pi_{\theta_t}(s), \quad (\mu_e, \sigma_e^2) \sim b_{t-1}(k),\quad s \in C_k
\label{eq:sac_updateb}
\end{equation}
At the same time, for any sample tuple $(s_i, a_i, r_i, s_i')$ at time step $t$, we use both $s_i$ and $s_i'$ to obtain the corresponding $(\mu_{\pi_{\theta_t}}(s_i), \sigma^2_{\pi_{\theta_t}}(s_i))$ and $(\mu_{\pi_{\theta_t}}(s_i'), \sigma^2_{\pi_{\theta_t}}(s_i'))$ for the actor and critic, respectively. This fusion method can then be formally defined as:
\begin{equation}
\begin{aligned}
    \mu_{\text{actor}}(s_i) &= (1 - \beta_t)\mu_{\pi_{\theta_t}}(s_i) + \beta_t\mu_b, \quad 
    \sigma^2_{\text{actor}}(s_i) = (1 - \beta_t)\sigma^2_{\pi_{\theta_t}}(s_i) + \beta_t\sigma^2_b,\\
    \mu_{\text{critic}}(s_i') &= (1 - \beta_t)\mu_{\pi_{\theta_t}}(s_i') + \beta_t\mu_b, \quad 
    \sigma^2_{\text{critic}}(s_i') = (1 - \beta_t)\sigma^2_{\pi_{\theta_t}}(s_i') + \beta_t\sigma^2_b, \\
    \text{where} \quad &(\mu_b, \sigma^2_b) \sim b_t(k), \quad s_i, s_i' \in C_k
\end{aligned}
\end{equation}
where $\beta_t \in [0,1]$ adaptively controls the relative weighting between task-driven and concept-informed signals, 
and $\mu_b, \sigma^2_b$ denote the currently stored conceptual experience in $b_t(k)$. This results in the conceptual belief-informed distribution for both the actor and critic:
\begin{equation}
    B_t(\cdot \mid s_i = \mathcal{N}(\mu_{\text{actor}}(s_i), \sigma^2_{\text{actor}}(s_i)), \quad
    B_t(\cdot \mid s_i') = \mathcal{N}(\mu_{\text{critic}}(s_i'), \sigma^2_{\text{critic}}(s_i')).
\label{eq:sac_B}
\end{equation}
Finally, we replace the policy $\pi_{\theta}$ in Eq.\ref{eq:sacupdate} with the integrated $B_t$ and perform the updates accordingly:
\begin{equation}
\begin{aligned}
\mathcal{L}_{\text{HI-SAC}_{critic}}(\phi_i) &= 
\mathbb{E}\Big[ \big(Q_{\phi_i}(s_i,a_i) - y_t \big)^2 \Big], \quad i=1,2, \\
\text{where} \quad
y_t & = r_i + \gamma \, \mathbb{E}_{a_i' \sim B_t}
\big[ Q_{\min}(s_i', a_i') - \alpha \log B_t(a_i' \mid s_i') \big], \\
\mathcal{L}_{\text{HI-SAC}_{actor}}(\theta) &= 
\mathbb{E}_{s_i \sim \mathcal{D}_t,\, a_i \sim B_t}
\big[ \alpha \log B_t(a_i \mid s_i) -  \min\{Q_{\phi_1}(s_i,a_i), \, Q_{\phi_2}(s_i,a_i)\} \big].
\label{eq:sac_updateparameter}
\end{aligned}
\end{equation}
By integrating policy learning with semantically grounded beliefs, HI-SAC enables agents to generalize across conceptually coherent behaviors. This fusion facilitates better sample reuse, long-term coherence, and more human-like decision-making. The pseudocodes are provided in Appendix \ref{al:sac}.

\section{Experiment}
\textbf{Experimental setup:} Evaluation is based on \textit{Feasible Cumulative Rewards}, where higher values indicate better performance, averaged over three seeds (123, 321, 666). The evaluation spans a wide range of environments, including Classic Control, Box2D \citep{catto2005iterative}, MetaDrive  \citep{li2022metadrive}, MuJoCo \citep{todorov2012mujoco}, and Atari \citep{bellemare2013arcade} domains. Conceptual clustering is simulated using clustering algorithms that group similar state-action pairs into latent categories. All methods employ identical hyperparameters and are implemented on the XuanCe benchmark suite \citep{liu2023xuance}.

\textbf{Evaluated methods:} For discrete action spaces, we compare HI-Q and HI-PPO with the following baselines: DQN \citep{mnih2013playing}, DDQN \citep{van2016deep}, DuelDQN \citep{wang2016dueling}, and PPO\citep{schulman2017proximal}, covering standard Q-value approximations, decoupled action evaluation, state-action advantage estimation, and clipped policy optimization. For continuous action spaces, HI-SAC is compared with A2C \citep{mnih2016asynchronous}, PPO, SAC \citep{haarnoja2018soft}, and DDPG \citep{lillicrap2015continuous}, representing common policy-gradient and actor-critic methods with entropy regularization or deterministic gradients.

\begin{table*}[htbp]
\centering
\caption{Average cumulative rewards of HI-RL variants and baselines across discrete and continuous action environments.}
\label{tab:experiment_results_combined}
\resizebox{1.0\textwidth}{!}{
\begin{tabular}{lccccc}
\toprule
\multicolumn{6}{c}{\textbf{HI-RL for DQN Variants}} \\
\midrule
\textbf{Environment/Method}        & \textbf{HI-DQN} & \textbf{PPO} & \textbf{DQN} & \textbf{Duel\_DQN} & \textbf{DDQN} \\
\midrule
\textbf{Classic Control - CartPole}     & \textbf{499.78 $\pm$ 0.22}    & 499.17 $\pm$ 0.83   & 478.44 $\pm$ 21.56  & 440.69 $\pm$ 59.31  & 396.51 $\pm$ 103.49 \\
\textbf{Classic Control - Acrobot}      & \textbf{-80.57 $\pm$ 17.48}   & -500.00 $\pm$ 0.00  & -87.19 $\pm$ 18.55  & -104.53 $\pm$ 54.19 & -100.77 $\pm$ 24.79 \\
\textbf{Box2d - CarRacing}  & \textbf{854.66 $\pm$ 45.35}   & 189.05 $\pm$ 56.48  & 830.78 $\pm$ 51.61  & -13.05 $\pm$ 24.66  & 766.16 $\pm$ 88.22  \\
\textbf{Box2d - LunarLander} & \textbf{232.73 $\pm$ 40.20}   & 204.95 $\pm$ 48.77  & 52.67 $\pm$ 192.08  & -58.97 $\pm$ 4.08   & 191.79 $\pm$ 69.16  \\
\textbf{MetaDrive - rXTSC}  & \textbf{189.22 $\pm$ 63.71}   & 156.74 $\pm$ 31.44  & 82.05 $\pm$ 82.84   & 39.50 $\pm$ 7.27    & 185.55 $\pm$ 107.80 \\
\textbf{MetaDrive - TOrSX}  & \textbf{159.39 $\pm$ 38.40}   & 149.97 $\pm$ 26.28  & 101.60 $\pm$ 13.72  & 69.16 $\pm$ 14.07   & 83.77 $\pm$ 22.37   \\
\textbf{MetaDrive - XTOC}   & \textbf{303.15 $\pm$ 50.89}   & 293.72 $\pm$ 66.42  & 170.73 $\pm$ 31.60  & 67.42 $\pm$ 6.29    & 170.73 $\pm$ 31.60  \\
\textbf{MetaDrive - XTSC}   & \textbf{233.91 $\pm$ 64.92}   & 191.50 $\pm$ 39.31  & 215.94 $\pm$ 205.74 & 63.47 $\pm$ 4.96    & 147.71 $\pm$ 92.55  \\
\textbf{MetaDrive - CYrXT}  & \textbf{97.99 $\pm$ 25.43}    & 97.83 $\pm$ -38.66  & 77.23 $\pm$ 47.94   & 9.12 $\pm$ 39.53    & 75.39 $\pm$ 49.99   \\
\textbf{MetaDrive - COrXSrT} & \textbf{117.90 $\pm$ 24.56}   & 89.27 $\pm$ 23.52   & 117.18 $\pm$ 30.28  & 53.01 $\pm$ 4.91    & 29.15 $\pm$ 16.26   \\
\textbf{MetaDrive - SrOYCtryS} & \textbf{130.27 $\pm$ 117.07} & 75.38 $\pm$ 8.12    & 105.01 $\pm$ 88.37  & 38.90 $\pm$ 0.39    & 100.72 $\pm$ 81.92  \\
\midrule
\multicolumn{6}{c}{\textbf{HI-RL for SAC Variants}} \\
\midrule
\textbf{Environment/Method}        & \textbf{HI-SAC} & \textbf{SAC} & \textbf{PPO} & \textbf{DDPG} & \textbf{A2C} \\
\midrule
\textbf{Box2d - BipedalWalker}  & \textbf{295.16 $\pm$ 99.64} & 285.71 $\pm$ 11.43 & -17.21 $\pm$ 45.45  & -34.58 $\pm$ 8.92  & -115.66 $\pm$ 1.95 \\
\textbf{Mujoco - Ant}            & \textbf{2862.15 $\pm$ 606.91} & 2386.54 $\pm$ 489.76 & 108.47 $\pm$ 14.97 & 2351.56 $\pm$ 147.15 & 1566.19 $\pm$ 346.25 \\
\textbf{Mujoco - Humanoid}       & \textbf{3248.46 $\pm$ 812.84} & 2090.07 $\pm$ 2233.68 & 52.35 $\pm$ 0.08   & 401.39 $\pm$ 84.60   & 179.26 $\pm$ 74.62 \\
\textbf{Mujoco - HumanoidStandup} & \textbf{132391.49 $\pm$ 606.23} & 121643.72 $\pm$ 25.53 & 112603.41 $\pm$ 65.06 & 69209.17 $\pm$ 14951.33 & 80250.37 $\pm$ 46.46 \\
\textbf{Mujoco - Reacher}        & \textbf{-3.96 $\pm$ 0.71}   & -4.65 $\pm$ 1.77  & -6.88 $\pm$ 0.08  & -5.73 $\pm$ 0.96  & -10.88 $\pm$ 0.12 \\
\textbf{Mujoco - HalfCheetah} & \textbf{10276.66 $\pm$ 2448.76} & 9678.01 $\pm$ 810.58 & 7378.66 $\pm$ 1951.02 & 3574.82 $\pm$ 2267.63 & 3043.32 $\pm$ 388.69 \\
\textbf{Mujoco - Hopper} & \textbf{3121.56 $\pm$ 573.84} & 2246.74 $\pm$ 657.82 & 1530.17 $\pm$ 1869.52 & 2338.46 $\pm$ 1075.83 & 520.53 $\pm$ 25.98 \\
\textbf{Mujoco - Walker2d} & \textbf{4444.48 $\pm$ 292.20} & 3382.66 $\pm$ 1177.36 & 992.81 $\pm$ 1799.20 & 3756.60 $\pm$ 840.68 & 733.50 $\pm$ 755.30 \\
\textbf{Mujoco - Pusher} & \textbf{-25.44 $\pm$ 6.16} & -31.76 $\pm$ 4.15 & -36.36 $\pm$ 0.82 & -45.50 $\pm$ 3.14 & -55.29 $\pm$ 1.65 \\
\textbf{Mujoco - InvertedPendulum} & \textbf{998.13 $\pm$ 1.87} & 860.78 $\pm$ 590.78 & 609.51 $\pm$ 4.51 & 973.82 $\pm$ 26.18 & 991.25 $\pm$ 116.64 \\
\textbf{Mujoco - InvertedDoublePendulum} & \textbf{9247.71 $\pm$ 103.30} & 8703.18 $\pm$ 644.18 & 126.87 $\pm$ 56.87 & 6444.11 $\pm$ 3857.15 & 7981.28 $\pm$ 1365.03 \\
\bottomrule
\end{tabular}
}
\end{table*}

\begin{table*}[ht]
\centering
\caption{Average cumulative rewards of HI-PPO, HI-TD3 and baselines across discrete and continuous action environments.}
\label{tab:experiment_results_combinedPPO}
\resizebox{1.0\textwidth}{!}{
\begin{tabular}{lccccc}
\toprule
\multicolumn{6}{c}{\textbf{HI-RL for PPO Variants}} \\
\midrule
\textbf{Method/Environment}        & \textbf{Atari - AirRaid} & \textbf{Atari - Amidar} & \textbf{Atari - Asteroids} & \textbf{Atari - Centipede} & \textbf{Atari - Zaxxon}\\
\midrule
\textbf{PPO} & 7210.01 $\pm$ 1594.32  & 917.56$ \pm$ 65.08 & 4190.79$\pm$ 928.38   & 4792.76$\pm$ 1244.33  & 15690.27$\pm$ 3486.71\\
\textbf{HI-PPO} & \textbf{9659.79 $\pm$ 2333.36} & \textbf{2302.55 $\pm$ 627.01} & \textbf{4419.23$\pm$ 1404.85} & \textbf{ 6002.09$\pm$ 1495.15} & \textbf{ 16663.71$\pm$5093.18}\\
\midrule
\multicolumn{6}{c}{\textbf{HI-RL for TD3 Variants}} \\
\midrule
\textbf{Method/Environment}        & \textbf{Box2d - BipedalWalker} & \textbf{Mujoco - Ant} & \textbf{Mujoco - Swimmer} & \textbf{Mujoco - HalfCheetah} & \textbf{Mujoco - Walker2d} \\
\midrule
\textbf{TD3} & 276.03 $\pm$ 42.42 &  5634.15$\pm$ 620.63& 50.50$\pm$ 1.45 &13194.89$\pm$ 755.84 & 4565.46$\pm$ 147.63\\
\textbf{HI-TD3} & \textbf{291.86 $\pm$ 23.45} & \textbf{6358.90$\pm$ 420.75} & \textbf{132.89$\pm$ 1.99} & \textbf{13706.98$\pm$399.64} & \textbf{6194.91$\pm$ 319.83}\\
\bottomrule
\end{tabular}
}
\end{table*}

\begin{figure*}[ht]
    \centering
    \includegraphics[width=1\textwidth]{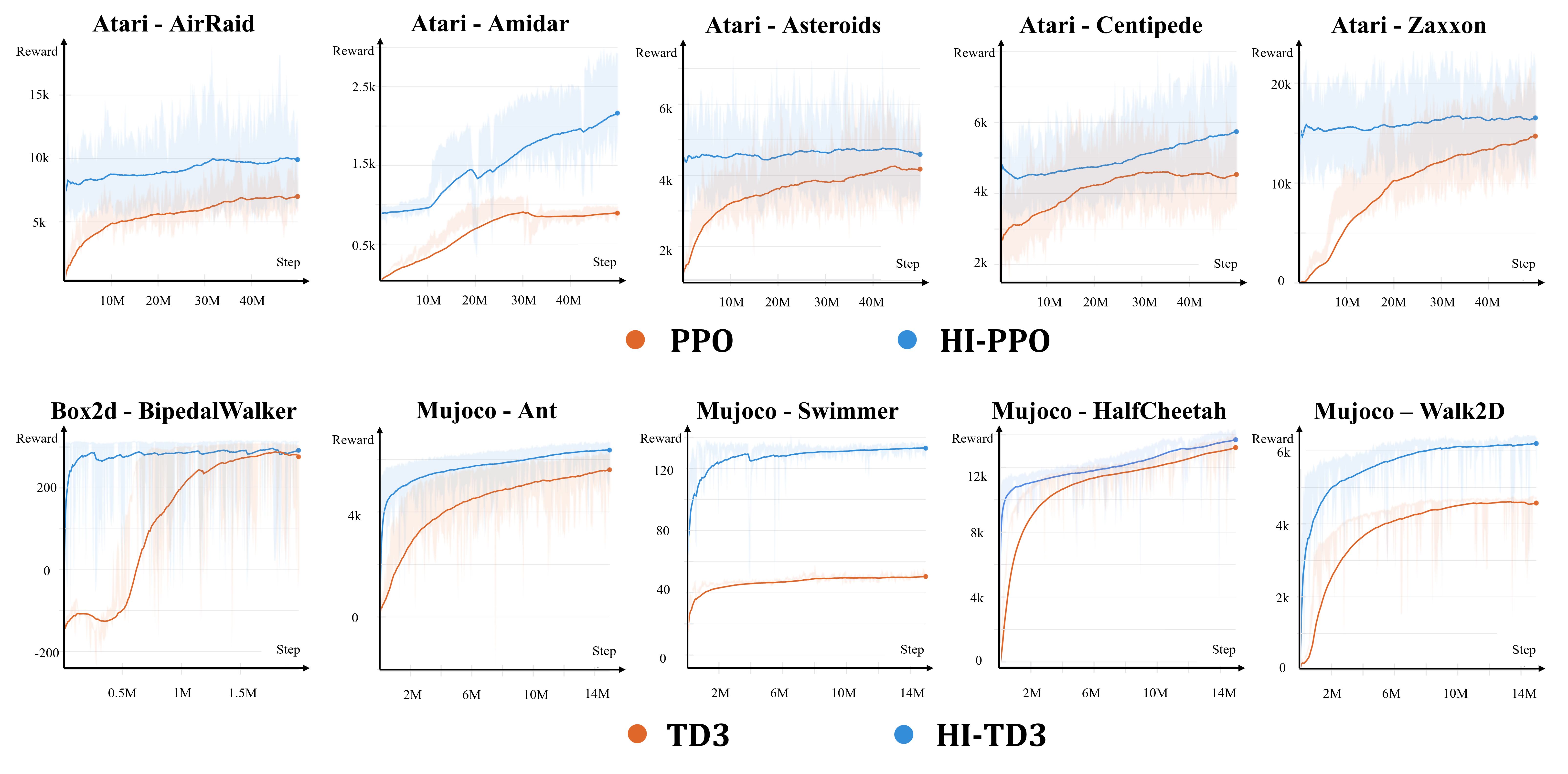}
    \caption{Learning curves comparing HI-PPO and PPO (Atari tasks) as well as HI-TD3 and TD3 (Mujoco and Box2D tasks). HI-RL variants demonstrate faster convergence, higher sample efficiency, and reduced variance across diverse environments.}
    \label{fig:learning_curves}
\end{figure*}

\subsection{Comparative Performance of HI-RL and Baselines}
To rigorously evaluate the HI-RL framework, we report results across a broad set of benchmark environments spanning both discrete and continuous action spaces (Table~\ref{tab:experiment_results_combined}, Table~\ref{tab:experiment_results_combinedPPO}). The tasks range from low-dimensional control (Classic Control, Box2D) to high-dimensional, perceptually rich domains (MetaDrive, MuJoCo), enabling a systematic assessment of generalization and sample efficiency under varying levels of complexity.

\textbf{Discrete Action Space:} As shown in Table~\ref{tab:experiment_results_combined}, HI-DQN (HI-Q) consistently outperforms baselines (DQN, DDQN, Dueling DQN, PPO) across diverse discrete-action tasks. In simple settings such as \textit{CartPole}, HI-DQN nearly reaches the performance ceiling with lower variance. In more complex tasks like \textit{Box2D-CarRacing} and MetaDrive, HI-DQN achieves the highest rewards across all sub-tasks, demonstrating robustness and adaptability. Even in intermediate (\textit{TOrSX}) and highly challenging scenarios (\textit{XTOC}), HI-DQN maintains clear advantages, highlighting the effectiveness of belief-guided abstraction for stable learning under increasing complexity.

\textbf{Continuous Action Space:} A similar trend is observed in continuous-control benchmarks (Table~\ref{tab:experiment_results_combined}). HI-SAC consistently outperforms SAC, PPO, and DDPG across both medium- and high-dimensional MuJoCo and Box2D tasks. In challenging domains such as \textit{Humanoid} and \textit{HumanoidStandup}, HI-SAC achieves substantially higher rewards with improved stability, while in locomotion tasks (\textit{HalfCheetah}, \textit{Walker2d}), it converges faster and produces more resilient policies. Overall, these results demonstrate that HI-SAC leverages belief-guided generalization to deliver reliable gains in environments requiring both precise control and long-horizon reasoning.

\subsection{Learning Dynamics with Experience-Driven Abstraction}
While the previous section demonstrates that HI-RL achieves superior final performance over baseline algorithms in both discrete and continuous action spaces, practical reinforcement learning often places greater emphasis on sample efficiency, training stability, and convergence speed than on post-convergence metrics. These factors are especially critical in resource-constrained or high-risk settings. To this end, we analyze the learning dynamics of HI-PPO vs. PPO and HI-TD3 vs. TD3 on Atari and MuJoCo (Fig.~\ref{fig:learning_curves}, Table~\ref{tab:experiment_results_combinedPPO}), illustrating how HI-RL leverages cognitive belief priors for faster exploration and structured abstraction for more stable optimization.

In high-dimensional visual environments such as Atari, HI-PPO consistently improves both convergence speed and final performance. For example, in \textit{Amidar}, HI-PPO surpasses $2000$ reward at $40$M steps, whereas PPO converges around $\sim900$. In more challenging tasks such as \textit{Asteroids} and \textit{Centipede}, HI-PPO not only learns faster but also exhibits reduced variance, indicating more stable policy updates. The progressive increase of $\beta_t$ enables HI-PPO to exploit conceptual priors early on and transition smoothly to task-specific fine-tuning, resulting in efficient and robust learning.

Similarly, in continuous control tasks, HI-TD3 achieves faster convergence, higher rewards, and greater stability compared to TD3. In simpler tasks such as \textit{BipedalWalker}, HI-TD3 converges more rapidly and attains comparable or better final performance. In more complex locomotion tasks including \textit{Ant}, \textit{Swimmer}, \textit{HalfCheetah}, and \textit{Walker2d}, HI-TD3 not only reaches higher asymptotic rewards but also produces smoother learning curves with lower variance. By contrast, TD3 often suffers from slower convergence and mid-training stagnation, underscoring the efficiency and robustness advantages of HI-TD3.

\section{Conclusion}  
We introduce Conceptual Belief-Informed Reinforcement Learning (\ours), a representation-level framework that organizes experiences into conceptual categories and integrates belief-guided fusion into policy learning. Moving beyond buffer replay and static policy libraries, \ours establishes a structured memory that supports abstraction, reuse, and generalization. Across Q-learning, PPO, TD3, and SAC, it consistently improves sample efficiency, final returns, and stability in both discrete and continuous domains. By achieving higher returns with fewer interactions and stabilizing updates, \ours also reduces computational cost, underscoring its potential for sustainable and resource-efficient training. More broadly, HI-RL illustrates how cognitive principles—conceptual abstraction and belief—can be operationalized to advance reinforcement learning, shifting the field from raw data manipulation toward structured, human-aligned inference. We view this as a step toward an “Era of Experience,” in which intelligence is grounded in the active organization of interaction history rather than rote prediction from data.

\newpage
\bibliography{iclr2026_conference}
\bibliographystyle{iclr2026_conference}

\newpage
\appendix
\section{Appendix}
\subsection{Conceptual Belief-Informed Twin Delayed Deep Deterministic Policy Gradient (HI-TD3)}
\label{al:td3_implement}
TD3 (Twin Delayed Deep Deterministic Policy Gradient)\citep{fujimoto2018addressing}, built upon DDPG\citep{silver2014deterministic}, mitigates Q-value overestimation and improves stability via twin Q-networks, delayed updates, and target policy smoothing. Here, we focus only on the actor update. Considering a sample batch $D_t =(s_i, a_i, r_i, s_i')$ at time step t, the actor policy update is defined as:
\begin{equation}
    \mathbb{E}_{s_i \sim D_t} \big[ Q_{\phi_{\min}}(s_i, \pi_{\theta}(s_i)) \big]
\end{equation}
where $Q_{\phi_{\min}}$ takes the smaller value of the two Q-networks, while $\pi_{\theta}$ denotes the policy that generates actions, with $a_i = \pi_{\theta}(s_i)$. Conceptual Belief-Informed TD3 (HI-TD3) applies the HI-RL fusion rule to deterministic policy gradients, refining the concept-based belief prior $b_t(k)$ for each conceptual category $C_k$ at time step t as: 
\begin{equation}
    \forall (\nabla,k) \in  \nabla \times [K]: \quad  b_t(k) \simeq \nabla_a Q_{\phi_{\min}}(s,a)
\end{equation}
where $b_t(k)$, a directional belief, denotes as $\nabla_a Q_{\min}(s,a)$, where the smaller Q-value in TD3 is employed to approximate the gradient serving as its representation. The recorded belief direction is updated using an exponential moving average with normalization:
\begin{equation}
    b_t(k)=\frac{ (1- \eta)b_{t-1}(k) + \eta \nabla_a Q_{\min}(s,a)}{\| (1- \eta)b_{t-1}(k) + \eta \nabla_a Q_{\min}(s,a)\|}, \quad a \sim \pi_{\theta}(s), \quad s \in C_k
\label{eq:td3_update}
\end{equation}
where $\eta$ is an exponential moving average constant and $b_{t-1}(k)$ denotes the previously stored directional belief. 

In the policy update of HI-TD3, we perform belief fusion updates only on the actor network. At each time step \textit{t} with sampled tuple $(s_i, a_i, r_i, s_i')$, the integrated directional belief information $B_t(k)$ is denotes as:
\begin{equation}
    B_t(k) = c \frac{(1 - \beta) \nabla_{a_i} Q_{\min}(s_i,a_i) + \beta b_t(k)}{\|(1 - \beta) \nabla_{a_i} Q_{\min}(s_i,a_i) + \beta b_t(k)\|}, \quad s_i \in C_k
\label{eq:td3_belief}
\end{equation}
where \textit{c} is a constant used to prevent excessive oscillations if $B_t(k)$ becomes too large. Differing from previous usage, $\beta$ is determined by directional similarity, computed as a dot product, and serves as the fusion coefficient: 
\begin{equation}
    \beta = \mathrm{clamp}\bigl(\sum_kb_t(k) \cdot \nabla_{a_i} Q_{\phi_{\min}}(s_i,a_i), 0 , 1 \bigr)
\label{eq:td3_beta}
\end{equation}
The directional fusion is performed by combining $B_t(k) $ as a perturbation with $a_i$:
\begin{equation}
    a_{\rm blend}=\mathrm{clamp}\bigl(a_i+B_t(k),-1,1\bigr)
\label{eq:td3_blend}
\end{equation}
The actor minimizes to update policy:
\begin{equation}
     \mathbb{E}_{s_i \sim D_t} \big[ Q_{\phi_{\min}}(s_i, a_{blend}) \big]
\label{eq:td3_fuse}
\end{equation}

Thus, HI-TD3 preserves the HI-RL fusion principle through the actor by blending task-driven gradients with conceptual priors, while the critic remains the standard TD3 update for stability. This makes HI-TD3 a deterministic yet framework-consistent instantiation of HI-RL. The pseudocodes are provided in Appendix \ref{al:td3}.

\newpage    
\subsection{Pseudo Code}
\subsubsection{Conceptual Belief-Informed Q-Learning (HI-Q) Algorithm}
\label{al:Q}
\begin{algorithm}[H]

    \caption{Conceptual Belief-Informed Q-learning (HI-Q) Algorithm}\label{algorithm:belief_q}
    \begin{algorithmic}[1]
        
        \State Initialization: learning rate $\alpha$, discount factor $\gamma$, running steps $T$, episodes $E$, replay buffer $\mathcal{B}$ and a set of K conceptual categories, denoted as  $\{\mathcal{C}_k\}^K_{k=1}$
        
        \For {each episode}
        \State Get initial state $s_0$ from the environment
        \For {each timestep t}
        
        \State Choose a random action $a_t$ with probability $\epsilon$ otherwise take  $a_t = \arg\max_{a} Q(s_t, a; \theta)$
        \State Execute $a_t$ to get reward $r(s_t,a_t)$, next state $s_{t+1}$
        \State Store $(s_t, a_t, r(s_t,a_t),s_{t+1} )$ into $\mathcal{B}$
        \State Identify the conceptual category $\mathcal{C}_k$ of $s_t$ through Nearest Neighbor
        \State Update the count of $a_t$ in $\mathcal{C}_k$ (cf.~\ref{eq:discrete-belief});
        
        \State Sample $N$ tuples from $\mathcal{B}$ to update $Q$ function:
        \State  \quad \quad Extract $b_t(a \mid \mathcal{C}_k(s_t))$ and integrate with rewards to estimate $B_t(a \mid s_{t+1})$ (cf.\ref{eq:Q-beta})
        \State \quad \quad $y^i_{s_t,a_t}=\mathbb{E}_{\mathcal{B}}\left[r(s_t,a_t)+\gamma\sum_{a} B_t(a \mid s_{t+1}) Q(s_{t+1}, a;\theta^-)| s_t,a_t\right]$ (cf.\ref{eq:q-target})
        \State \quad \quad $Loss = \mathbb{E}_{\mathcal{B}}\left[(y_{s_t,a_t}^{i}-Q(s_t,a_t;\theta))^{2}\right]$
        \State Reset target network after a few updates: align target Q parameters: $\theta^{-} = \theta$;
        \EndFor
        \EndFor
    \end{algorithmic}
\end{algorithm}
\newpage
\subsubsection{Conceptual Belief-Informed Soft Actor-Critic (HI-SAC) Algorithm}
\label{al:sac}
\begin{algorithm}[H]
\caption{Conceptual Belief-Informed Soft Actor-Critic}
\begin{algorithmic}[1]
\State Initialize two critic parameters $\phi_1$, $\phi_2$ and actor parameters $\theta$, Conceptual categories $\{C_k\}_{k=1}^N$, category belief parameters $b_{t=0}(k) = \{\mu_k, \sigma_k^2\}_{k=1}^N$
\For{each time step t}
    \State Sample $a_t \sim \pi_{\theta}(\cdot \mid s_t)$
    \State Transition to $s_{t+1} \sim p(s_{t+1} \mid s_t, a_t)$
    \State Store transition in replay buffer: $\mathcal{B} \gets \mathcal{B} \cup \{(s_t, a_t, r(s_t, a_t), s_{t+1})\}$
    \For{training step }
        \State Sampled $\{s_i, a_i, r_i, s_i' \}\gets \mathcal{B}$ 
        \State Identify category $C_k$ for $s_i$ through Euclidean distance
        \State Compute $B_t(si) $ and $B_t(si')$, $s_i, s_i' \in C_K$ (cf.~\ref{eq:sac_B})
        \State Update category belief parameters $b_t(k)$ (cf.~\ref{eq:sac_updateb})
    \EndFor
    \For{each gradient step (cf.\ref{eq:sac_updateparameter})}
        \State Compute target:
    \[
        y_i = r_i + \gamma \, \mathbb{E}_{a'_i \sim B_t(\cdot|s'_i)} 
        \Big[ Q_{\min}(s'_i,a'_i) - \alpha \log B_t(a'_i|s'_i) \Big].
    \]

    \State Update critics $(i=1,2)$:
    \[
        L_{\text{CBISAC}}^{\text{critic}}(\phi_i) 
        = \mathbb{E}_{(s_i,a_i)\sim D_t} 
        \Big[ \big(Q_{\phi_i}(s_i,a_i) - y_i \big)^2 \Big],
    \]
    \[
        \phi_i \leftarrow \phi_i - \eta_\phi \nabla_{\phi_i} L_{\text{CBISAC}}^{\text{critic}}(\phi_i).
    \]

    \State Update actor:
    \[
        L_{\text{CBISAC}}^{\text{actor}}(\theta) 
        = \mathbb{E}_{s_i \sim D_t,\, a_i \sim B_t}
        \Big[ \alpha \log B_t(a_i|s_i) - Q_{\min}(s_i,a_i) \Big],
    \]
    \[
        \theta \leftarrow \theta - \eta_\theta \nabla_\theta L_{\text{CBISAC}}^{\text{actor}}(\theta).
    \]

    \State Update temperature:
    \[
        L(\alpha) = \mathbb{E}_{s_i,a_i\sim B_t} 
        \big[ -\alpha \, (\log B_t(a_i|s_i) + \mathcal{H}_{\text{target}}) \big],
    \]
    \[
        \alpha \leftarrow \alpha - \eta_\alpha \nabla_\alpha L(\alpha).
    \]

    \State Soft update target network:
    \[
        \bar{\phi} \gets \tau \phi + (1-\tau)\bar{\phi}.
    \]
    \State Update $b_t(k)$  (cf.\ref{eq:sac_updateb}) 
    \EndFor
\EndFor
\end{algorithmic}
\end{algorithm}

\newpage
\subsubsection{Conceptual Belief-Informed Proximal Policy Optimization (HIPPO) Algorithm}
\label{al:ppo}
\begin{algorithm}[H]
\caption{Conceptual Belief-Informed Proximal Policy Optimization}
\begin{algorithmic}[1]
\State Initialize policy parameters $\theta$ and value function parameters $\phi$, conceptual categories $\{C_k\}_{k=1}^N$
\For{each iteration}
    \For{each environment step t}
        \State Collect set of trajectories $D_k = \{\tau_i\}$ by running $\pi_k = \pi(\theta_k)$
        \State Sample $a_t$ and Transition to get $s_{t+1}$
        \State Compute rewards-to-go $r(s_t,a_t)$.
        \State Compute advantage estimation $A_t$ based on current value function $V_{\phi_k}$
        \State Store transition in replay buffer: $\mathcal{B} \gets \mathcal{B} \cup \{(s_t, a_t, r(s_t, a_t), s_{t+1},A_t)\}$
    \EndFor
    \For{each gradient step}
        \State Sampled $\{s_i, a_i, r_i, s_i' \}\gets \mathcal{B}$ 
        \State Identify category $C_k$ for $s_i$ through Euclidean distance
        \State Compute $B_t(k) = (1 - \beta_t) \pi_{\theta}(a_i \mid s_i) + \beta_tb_t(k), \quad s_i \in C_k$ (cf.\ref{eq:cbdppo_beta})
        \State Update the policy by maximizing the PPO-Clip objective (cf.\ref{eq:cbdppo_loss}):\\
        \[
        \theta_{k+1} \;=\; 
        \arg\max_{\theta}\;
        \mathbb{E}_{(s_i,a_i)\sim\pi_{\theta_{\mathrm{old}}}}\Bigg[
        \min\!\Bigg(
        \frac{B_t(a_i \mid s_i)}{\pi_{\theta_{\mathrm{old}}}(a_i \mid s_i)}\,A_i,\;
        \operatorname{clip}\!\Big(
        \frac{B_t(a_i \mid s_i)}{\pi_{\theta_{\mathrm{old}}}(a_i \mid s_i)},\,
        1-\epsilon,\,1+\epsilon
        \Big) A_i
        \Bigg)
        \Bigg].
        \]
       
        \State Fit value function by regression on mean-squared error:\\
        \[
        \phi_{k+1} \;=\; \arg\min_{\phi}\;\mathbb{E}_{(s_i,a_i)\sim\pi_{\theta_{\mathrm{old}}}}
        \Big[
        \big( V_\phi(s_i) - r(s_i, a_i) \big)^2
        \Big].
        \]
    \State Update $b_t(k)$ (cf.\ref{eq:discrete-belief})
    \EndFor
\EndFor
\end{algorithmic}
\end{algorithm}

\newpage
\subsubsection{Conceptual Belief-Informed Twin Delayed Deep Deterministic (HI-TD3) Algorithm}
\label{al:td3}
\begin{algorithm}[H]
\caption{Conceptual Belief-Informed Twin Delayed Deep Deterministic}
\begin{algorithmic}[1]
\State Initialize actor $\pi_{\theta}(s)$, critics $Q_{\phi_1}(s,a), Q_{\phi_2}(s,a)$, target networks $\theta' \leftarrow \theta$, $\phi_1' \leftarrow \phi_1$, $\phi_2' \leftarrow \phi_2$, Conceptual categories $\{C_k\}_{k=1}^N$, discount factor $\gamma$ 
\For{each iteration}
    \For{each environment step t}
        \State Sample $a_t \;=\; \pi_{\theta}(s_t) + \epsilon, 
        \quad \epsilon \sim \mathcal{N}(0, \sigma^{2} I)$
        
        \State Transition to $s_{t+1} \sim p(s_{t+1} \mid s_t, a_t)$
        \State Store transition in replay buffer: $\mathcal{B} \gets \mathcal{B} \cup \{(s_t, a_t, r(s_t, a_t), s_{t+1})\}$
    \EndFor
    \For{each gradient step}
       \State Sample minibatch $D_t = \{(s_i, a_i, r_i, s'_i)\}$ from replay buffer
        \State \textbf{Critic update (TD3)}
        \State \hspace{1em} Compute target action with smoothing: $a_i' = \pi_{\theta'}(s'_i) + \epsilon, \quad \epsilon \sim \mathcal{N}(0,\sigma^2I)$

        \State \hspace{1em} Compute temporal difference target: $y_i = r_i + \gamma \cdot \min\big(Q_{\phi_1'}(s'_i,a_i'), Q_{\phi_2'}(s'_i,a_i')\big)$
    
        \State \hspace{1em} Update critics by minimizing loss: $L = \frac{1}{N}\sum_i \big(y_i - Q_{\phi_j}(s_i,a_i)\big)^2, \quad j=1,2$
    
        \State \textbf{Actor update with fusion}   
        \State \hspace{1em} Compute gradient direction: $g_t = \nabla_a Q_{\min}(s_i, a_i)$
        \State \hspace{1em}  Compute fusion coefficient $\beta$ (cf.\ref{eq:td3_beta}) 
        \State \hspace{1em} Integrate belief: $B_t = c \cdot \frac{(1-\beta) g_t + \beta b_t(k)}{\|(1-\beta) g_t + \beta b_t(k)\|}$ (cf.\ref{eq:td3_fuse})
        \State \hspace{1em} Blend action: $a_{\mathrm{blend}} = \mathrm{clamp}(\pi_{\theta}(s_i) + B_t, -1, 1)$(cf.\ref{eq:td3_blend})

        \State \hspace{1em} Update actor by minimizing: $J(\theta) = -\frac{1}{N}\sum_i Q_{\min}(s_i, a_{\mathrm{blend}}) $(cf.\ref{eq:td3_update})
    
        \State \textbf{Target network updates}
        \[
            \phi' \leftarrow \tau \phi + (1-\tau)\phi', 
            \quad \theta' \leftarrow \tau \theta + (1-\tau)\theta'
        \]
        \State Update $b_t(k)$ (cf.\ref{eq:td3_belief})
    \EndFor
\EndFor
\end{algorithmic}
\end{algorithm}

\subsection{Smoothed Bellman Operator}

To reflect cognitive properties of uncertainty-aware decision-making in reinforcement learning, we revise the classical Bellman operator, which updates values deterministically:
\begin{equation}
\mathcal{T}Q(s_t,a_t) = r_t + \gamma \max_{a \in \mathcal{A}} Q_t(s_{t+1},a).    
\end{equation}

Here, $\mathcal{T}$ is the classical Bellman operator, $s_t \in \mathcal{S}$ denotes the current state, $a_t \in \mathcal{A}$ the chosen action, $r_t \in \mathbb{R}$ the immediate reward, $\gamma \in (0,1)$ the discount factor, $\mathcal{A}$ the action space, and $Q_t(s,a)$ the action–value function at iteration $t$. The $\max_{a \in \mathcal{A}}$ term represents greedy action selection, i.e., propagating value based on the action with the highest estimated return.  

\begin{equation}
\mathcal{T}_{\text{Smoothed}} Q(s_t,a_t) = r_t + \gamma \sum_{a \in \mathcal{A}} q_t(a \mid s_{t+1}) Q_t(s_{t+1},a),    
\end{equation}

where $\mathcal{T}_{\text{Smoothed}}$ denotes the \emph{Smoothed Bellman Operator}, which replaces the hard maximization with a belief-weighted expectation. Here, $q_t(a \mid s_{t+1})$ is the action-preference distribution at state $s_{t+1}$, e.g., a softmax distribution over $Q_t(s_{t+1},a)$ or the belief-preference distribution in HI-RL. Unlike the deterministic $\max$, this formulation propagates value in a probabilistic, uncertainty-aware manner, balancing task-driven estimates with belief-informed priors.

This smoothing relaxes the deterministic backup, enabling value propagation to account for uncertainty and preference variability. The Smoothed Bellman Operator thus provides a unified, differentiable mechanism for propagating reward uncertainty. In Section~5, we illustrate how Smoothed Bellman Operator integrates with different policy learning paradigms (Q-learning, SAC, PPO). Formal instantiations such as softmax smoothing, clipped interpolation, and Bayesian fusion are provided in next, along with a convergence proof and a Jensen-type inequality.

\begin{lemma}[Jensen's Inequality for Q-values]
Consider an MDP with state \(s_{t+1}\) and actions \(a\), along with Q-value estimates \( Q_t(s_{t+1}, a) \). Let \(q_t(a \mid s_{t+1})\) denote the probability of selecting action \(a\) in state \(s_{t+1}\). By Jensen's inequality:
\begin{equation}
\begin{aligned}
\gamma \sum_{s_{t+1}} P(s_{t+1} \mid s_t, a_t) 
\sum_{a'} q_t(a \mid s_{t+1}) Q_t(s_{t+1}, a) 
&\leq \\
\gamma \sum_{s_{t+1}} P(s_{t+1} \mid s_t, a_t) 
\max_{a} Q_t(s_{t+1}, a),
\end{aligned}
\end{equation}
\label{lemma:jensen}
\end{lemma}

Lemma~\ref{lemma:jensen} establishes that the Smoothed Bellman Operator provides a conservative backup: replacing the hard maximization with a belief-weighted expectation yields an update that forms a lower bound on the classical Bellman backup, thereby stabilizing value propagation under uncertainty.

\begin{lemma}[Convergence of Smoothed Bellman Operator]
Let \(\{Q_t\}\) be the sequence generated by iteratively applying \( \mathcal{T}_{\text{Smoothed}} \). Under the condition:
\begin{equation}
\lim_{t \to \infty} \max_{a} q_t(a \mid s_{t+1}) = 1,
\end{equation}
for the optimal action, \(Q_t\) converges to the optimal \(Q^*\) as \(t \to \infty\). See Appendix D for a detailed proof.
\label{lemma:convergence2}
\end{lemma}

Lemma~\ref{lemma:convergence2} complements this by showing that if the action-preference distribution $q_t(\cdot \mid s_{t+1})$ asymptotically collapses onto the optimal action, then iterative application of the Smoothed Bellman Operator converges to the optimal value function $Q^*$. Together, these results establish that the Smoothed Bellman Operator not only smooths value backups for improved robustness, but also preserves the fundamental convergence guarantees of classical reinforcement learning. The full proof of Lemma~\ref{lemma:convergence2} is provided in next subsection.

To instantiate the Smoothed Bellman Operator in practice, different smoothing strategies can be employed to construct the action-preference distribution $b_t$. These strategies determine how strongly the update deviates from hard maximization and how uncertainty is incorporated. Representative examples are summarized in Table~\ref{table:smoothingstrategy}.

\begin{table}[h]
\centering
\setlength{\tabcolsep}{10mm}{
\begin{tabular}{|c|c|}
\hline
\textbf{Strategy} & \textbf{Formula} \\
\hline
Softmax & 
$q_t = \frac{e^{Q(s,a)}}{\sum_b e^{Q(s,b)}}$ \\
\hline
Clipped Max & 
\begin{minipage}{0.4\textwidth}
\[
q_t = \begin{cases} 
1 - \tau, & \text{if } a = a^* \\
\frac{\tau}{A - 1}, & \text{if } a \neq a^*
\end{cases}
\]
\end{minipage} \\
\hline
Clipped Softmax & 
\begin{minipage}{0.4\textwidth}
\[
q_t = \begin{cases} 
\frac{e^{\beta Q(s,a)}}{\sum_{b \in I} e^{\beta Q(s,b)}}, & \text{if } a \in I \\
0, & \text{if } a \notin I
\end{cases}
\]
\end{minipage} \\
\hline
\end{tabular}}
\caption{Smoothing strategies with respective formulas}
\label{table:smoothingstrategy}
\end{table}

\subsubsection{Convergence Proof}
We outline a proof that builds upon the following result \citep{singh2000convergence,barber2023smoothed} and follows the framework provided in \citep{melo2001convergence}:

\begin{theorem}
The random process \(\{\Delta_t\}\) taking value in \(\mathbb{R}\) and defined as
\begin{equation}
\Delta_{t+1}(x) = (1 - \alpha_t(x)) \Delta_t(x) + \alpha_t(x) F_t(x)
\end{equation}
converges to 0 with probability 1 under the following assumptions:
\begin{itemize}
    \item \(0 \leq \alpha_t \leq 1\), \(\sum_t \alpha_t(x) = \infty\), \(\sum_t \alpha_t^2(x) < \infty\);
    \item \(\mathbb{E}[\|F_t(x)\|_W] \leq \kappa \|\Delta_t\|_W + c_t\), \(\kappa \in [0, 1)\) and \(c_t \to 0\) with probability 1;
    \item \(\text{var}(F_t(x)) \leq C(1 + \|\Delta_t\|_W)^2\), \(C > 0\)
\end{itemize}
where \(\|\Delta_t\|_W\) denotes a weighted max norm.
\end{theorem}

We are interested in the convergence of \(Q_t\) towards the optimal value \(Q_*\) and therefore define
\begin{equation}
\Delta_t = Q_t(s_t, a_t) - Q_*(s_t, a_t)
\end{equation}
It is convenient to write the smoothed update as
\begin{equation}
Q_{t+1}(s_t, a_t) = Q_t(s_t, a_t) + \alpha_t(s_t, a_t)\left( r_t + \gamma \left\langle Q(s_{t+1}, a) \right\rangle_a - Q_t(s_t, a_t)\right)
\end{equation}
where \(\langle f(x) \rangle_x\) means the expectation of the function \(f(x)\) with respect to the distribution of \(x\). Using the smoothed update, we can write
\begin{equation}
\Delta_{t+1}(s_t, a_t) = Q_{t+1}(s_t, a_t) - Q_*(s_t, a_t)
\end{equation}
\begin{equation}
= (1 - \alpha_t)\Delta_t + \alpha_t\left( r_t + \gamma \langle Q(s_{t+1}, a)\rangle_a - Q_*(s_t, a_t) \right)
\end{equation}
In terms of Theorem 1, we therefore define
\begin{equation}
F_t = r_t + \gamma \sum_a q_t(a|s_{t+1}) Q_t(s_{t+1}, a) - Q_*(s_t, a_t)
\end{equation}

\begin{proof}
For convergence, we need to verify the conditions of Theorem 1.

\textbf{Step 1: Verify Step-Size Conditions}

We assume that the learning rates \(\alpha_t(s_t, a_t)\) satisfy:
\begin{itemize}
    \item \(0 < \alpha_t(s_t, a_t) \leq 1\),
    \item \(\sum_{t} \alpha_t(s_t, a_t) = \infty\),
    \item \(\sum_{t} \alpha_t^2(s_t, a_t) < \infty\).
\end{itemize}
An example is \(\alpha_t(s_t, a_t) = \frac{1}{N_t(s_t, a_t)}\), where \(N_t(s_t, a_t)\) is the visitation count of \((s_t, a_t)\).

\textbf{Step 2: Establish Boundedness of \(Q_t\)}

Since the rewards \( r_t \) are bounded (\( |r_t| \leq R_{\text{max}} \)) and the discount factor \( 0 < \gamma < 1 \), we can show that \( Q_t \) remains bounded independently of the convergence of \( \Delta_t \).

\textit{Define the Bound \( Q_{\text{max}} \):}

We define
\begin{equation}
Q_{\text{max}} = \frac{R_{\text{max}}}{1 - \gamma}.
\end{equation}
This is the maximum possible value of the Q-function given the bounded rewards and discount factor.

\textit{Derivation of \( Q_{\text{max}} \):}

The Q-function \( Q(s, a) \) represents the expected cumulative discounted reward when starting from state \( s \) and taking action \( a \):
\begin{equation}
Q(s, a) = \mathbb{E}\left[ \sum_{k=0}^\infty \gamma^k r_{t+k} \,\Big|\, s_t = s, a_t = a \right],
\end{equation}
where \( r_{t+k} \) is the reward received at time \( t + k \), and \( \gamma \) is the discount factor.

Assuming that at each time step, the agent receives the maximum possible reward \( R_{\text{max}} \), the maximum possible Q-value is:
\begin{equation}
Q_{\text{max}} = \sum_{k=0}^\infty \gamma^k R_{\text{max}} = R_{\text{max}} \sum_{k=0}^\infty \gamma^k.
\end{equation}

Since \( 0 < \gamma < 1 \), the infinite sum \( \sum_{k=0}^\infty \gamma^k \) is a geometric series that sums to:
\begin{equation}
\sum_{k=0}^\infty \gamma^k = \frac{1}{1 - \gamma}.
\end{equation}

Therefore, we have:
\begin{equation}
Q_{\text{max}} = R_{\text{max}} \times \frac{1}{1 - \gamma} = \frac{R_{\text{max}}}{1 - \gamma}.
\end{equation}

Thus, \( Q_{\text{max}} = \dfrac{R_{\text{max}}}{1 - \gamma} \) is the maximum possible value of the Q-function in any state-action pair.

\textit{Base Case:} Let \( Q_0(s, a) \) be initialized such that \( |Q_0(s, a)| \leq Q_{\text{max}} \) for all \( s, a \).

\textit{Inductive Step:} Assume \( |Q_t(s, a)| \leq Q_{\text{max}} \) for all \( s, a \). We need to show that \( |Q_{t+1}(s_t, a_t)| \leq Q_{\text{max}} \).

From the update equation:
\begin{equation}
Q_{t+1}(s_t, a_t) = Q_t(s_t, a_t) + \alpha_t(s_t, a_t)\left( r_t + \gamma \left\langle Q_t(s_{t+1}, a) \right\rangle_a - Q_t(s_t, a_t) \right).
\end{equation}

Simplifying:
\begin{equation}
Q_{t+1}(s_t, a_t) = (1 - \alpha_t(s_t, a_t)) Q_t(s_t, a_t) + \alpha_t(s_t, a_t)\left( r_t + \gamma \left\langle Q_t(s_{t+1}, a) \right\rangle_a \right).
\end{equation}

Taking absolute values:
\begin{equation}
|Q_{t+1}(s_t, a_t)| \leq (1 - \alpha_t) |Q_t(s_t, a_t)| + \alpha_t \left( |r_t| + \gamma \left| \left\langle Q_t(s_{t+1}, a) \right\rangle_a \right| \right).
\end{equation}

Using the inductive hypothesis and boundedness:
\begin{equation}
|Q_t(s_t, a_t)| \leq Q_{\text{max}}, \quad \left| \left\langle Q_t(s_{t+1}, a) \right\rangle_a \right| \leq Q_{\text{max}},
\end{equation}
and \( |r_t| \leq R_{\text{max}} \). Therefore:
\begin{equation}
|Q_{t+1}(s_t, a_t)| \leq (1 - \alpha_t) Q_{\text{max}} + \alpha_t \left( R_{\text{max}} + \gamma Q_{\text{max}} \right).
\end{equation}

Simplify:
\begin{align}
|Q_{t+1}(s_t, a_t)| &\leq Q_{\text{max}} - \alpha_t Q_{\text{max}} + \alpha_t \left( R_{\text{max}} + \gamma Q_{\text{max}} \right) \\
&= Q_{\text{max}} + \alpha_t \left( R_{\text{max}} - (1 - \gamma) Q_{\text{max}} \right).
\end{align}

Since \( Q_{\text{max}} = \dfrac{R_{\text{max}}}{1 - \gamma} \), we have \( (1 - \gamma) Q_{\text{max}} = R_{\text{max}} \). Substituting back:
\begin{equation}
|Q_{t+1}(s_t, a_t)| \leq Q_{\text{max}} + \alpha_t \left( R_{\text{max}} - R_{\text{max}} \right) = Q_{\text{max}}.
\end{equation}

Thus,
\begin{equation}
|Q_{t+1}(s_t, a_t)| \leq Q_{\text{max}}.
\end{equation}

Therefore, by induction, \( Q_t \) remains bounded for all \( t \), independently of \( \Delta_t \).

\textbf{Step 3: Verify Mean Condition}

We can write
\begin{equation}
\frac{1}{\gamma} \mathbb{E}[F_t] = \mathbb{E}_{p_{\mathcal{T}}}[ G_t ],
\end{equation}
where
\begin{equation}
G_t = \sum_a q_t(a|s_{t+1}) Q_t(s_{t+1}, a) - \max_a Q_*(s_{t+1}, a).
\end{equation}
We can form the bound
\begin{equation}
\left\| \frac{1}{\gamma} \mathbb{E}[F_t] \right\|_\infty = \left\| \mathbb{E}[ G_t ] \right\|_\infty \leq \| G_t \|_\infty,
\end{equation}
which means that if we can bound \(\|G_t\|_\infty\) appropriately, the mean criterion will be satisfied.

Assuming that \(b_t\) places \((1 - \delta_t)\) mass on the maximal action \(a^* = \arg\max_a Q_t(s_{t+1}, a)\), we can write
\begin{align}
G_t &= \sum_a q_t(a|s_{t+1}) Q_t(s_{t+1}, a) - \max_a Q_*(s_{t+1}, a) \\
&= (1 - \delta_t) Q_t(s_{t+1}, a^*) + \delta_t \sum_{c \ne a^*} \tilde{q}_t(c|s_{t+1}) Q_t(s_{t+1}, c) - \max_a Q_*(s_{t+1}, a),
\end{align}
where \(\tilde{b}_t(c|s_{t+1}) = \frac{b_t(c|s_{t+1})}{\delta_t}\) for \(c \ne a^*\).

We can then write
\begin{align}
G_t &= Q_t(s_{t+1}, a^*) - \max_a Q_*(s_{t+1}, a) + \delta_t \left( \sum_{c \ne a^*} \tilde{b}_t(c|s_{t+1}) [ Q_t(s_{t+1}, c) - Q_t(s_{t+1}, a^*) ] \right).
\end{align}
Since \( Q_t(s_{t+1}, a^*) \geq Q_t(s_{t+1}, c) \) for all \( c \), the terms inside the brackets are non-positive. Therefore,
\begin{align}
G_t &\leq Q_t(s_{t+1}, a^*) - \max_a Q_*(s_{t+1}, a).
\end{align}
Now, we have
\begin{align}
Q_t(s_{t+1}, a^*) - \max_a Q_*(s_{t+1}, a) &= [ Q_t(s_{t+1}, a^*) - Q_*(s_{t+1}, a^*) ] + [ Q_*(s_{t+1}, a^*) - \max_a Q_*(s_{t+1}, a) ] \\
&\leq \Delta_t(s_{t+1}, a^*).
\end{align}
Thus,
\begin{equation}
G_t \leq \Delta_t(s_{t+1}, a^*).
\end{equation}
Therefore,
\begin{equation}
\| G_t \|_\infty \leq \| \Delta_t \|_\infty.
\end{equation}
Additionally, the term involving \(\delta_t\) contributes an additional \( c_t \), which is bounded due to the boundedness of \(Q_t\) and \(\delta_t \to 0\). Thus, the mean condition becomes
\begin{equation}
\left\| \mathbb{E}[ F_t ] \right\|_\infty \leq \gamma \| \Delta_t \|_\infty + c_t,
\end{equation}
with \( c_t \to 0 \) as \( \delta_t \to 0 \).

Since \( \gamma < 1 \), the mean condition is satisfied with \( \kappa = \gamma \) and \( c_t \to 0 \).

\textbf{Step 4: Verify Variance Condition}

Since the rewards \( r_t \) are bounded and we have established that \( Q_t \) is bounded independently, \( F_t \) is also bounded.

We can write:
\begin{align}
\Delta F_t &= F_t - \mathbb{E}[ F_t ] \\
&= ( r_t - \mathbb{E}[ r_t | s_t, a_t ] ) + \gamma \left( \sum_a q_t(a|s_{t+1}) Q_t(s_{t+1}, a) - \mathbb{E}_{s_{t+1}} \left[ \sum_a q_t(a|s_{t+1}) Q_t(s_{t+1}, a) \right] \right ).
\end{align}

We can bound the variance using
\begin{equation}
\text{Var}(F_t) = \mathbb{E}\left[ (\Delta F_t)^2 \mid \mathcal{F}_t \right] \leq \| \Delta F_t \|_\infty^2.
\end{equation}
Using the triangle inequality,
\begin{align}
\| \Delta F_t \|_\infty &\leq \| \Delta r_t \|_\infty + \gamma \left\| \sum_a q_t(a|s_{t+1}) Q_t(s_{t+1}, a) - \mathbb{E}_{s_{t+1}} \left[ \sum_a q_t(a|s_{t+1}) Q_t(s_{t+1}, a) \right] \right\|_\infty \\
&\leq \| \Delta r_t \|_\infty + \gamma \left\| Q_t(s_{t+1}, a) - \mathbb{E}_{s_{t+1}} [ Q_t(s_{t+1}, a) ] \right\|_\infty.
\end{align}

Since \( Q_t \) is bounded, there exists a constant \( B \) such that
\begin{equation}
\| Q_t(s_{t+1}, a) - \mathbb{E}_{s_{t+1}} [ Q_t(s_{t+1}, a) ] \|_\infty \leq 2 Q_{\text{max}} = B.
\end{equation}
Therefore,
\begin{equation}
\| \Delta F_t \|_\infty \leq \| \Delta r_t \|_\infty + \gamma B.
\end{equation}
Since \( r_t \) is bounded, \( \| \Delta r_t \|_\infty \leq 2 R_{\text{max}} \).

Thus,
\begin{equation}
\| \Delta F_t \|_\infty \leq 2 R_{\text{max}} + \gamma B.
\end{equation}
Therefore, the variance is bounded, and there exists a constant \( C > 0 \) such that
\begin{equation}
\text{Var}(F_t) \leq C ( 1 + \| \Delta_t \|_\infty )^2.
\end{equation}

\textbf{Step 5: Conclusion}

All the conditions of Theorem 1 are satisfied:

\begin{itemize}
    \item \textbf{Step-Size Conditions:} Verified in Step 1.
    \item \textbf{Mean Condition:} Verified in Step 3, with \( \kappa = \gamma < 1 \) and \( c_t \to 0 \).
    \item \textbf{Variance Condition:} Verified in Step 4.
\end{itemize}

Therefore, \(\Delta_t \to 0\) with probability 1, implying that \( Q_t \to Q_* \) with probability 1.

\end{proof}

\newpage
\subsection{Experiment Setting}
\subsubsection{Classic Control and Box 2D Environment}
\begin{figure}[htbp]
    \centering
    \includegraphics[width=0.15\textwidth]{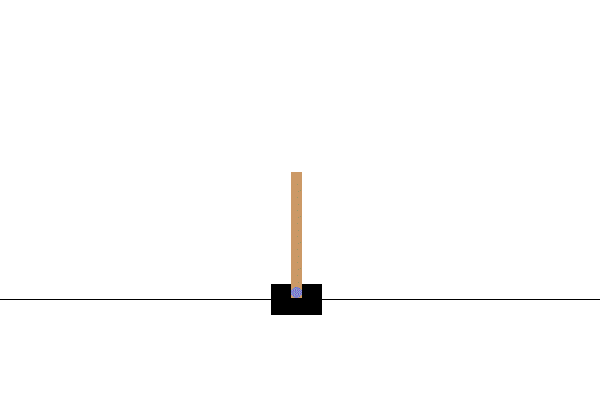}
    \includegraphics[width=0.15\textwidth]{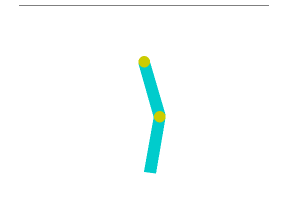}
    \includegraphics[width=0.15\textwidth]{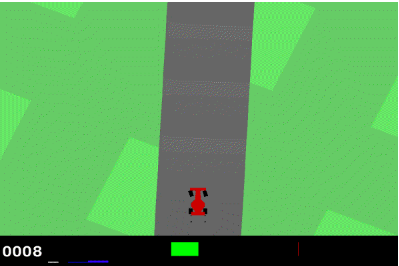}
    \includegraphics[width=0.15\textwidth]{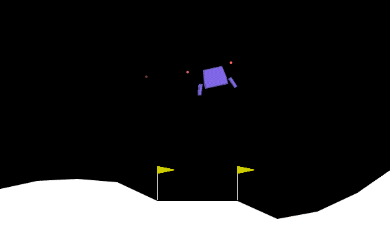}
    \includegraphics[width=0.15\textwidth]{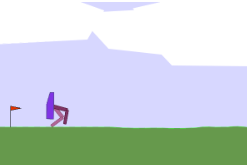}
    \caption{Cartpole, Acrobot, CarRacing, Lunar Lander and Bipedal Walker .}
    \label{fig:box2d_env}
\end{figure}

\begin{enumerate}
    \item Cartpole: a pole is attached by an unactuated joint to a cart, which moves along a frictionless track. The pendulum is placed upright on the cart and the goal is to balance the pole by applying forces in the left and right direction on the cart.
    \item Acrobot: a two-link pendulum system with only the second joint actuated. The task is to swing the lower link to a sufficient height in order to raise the tip of the pendulum above a target height. The environment challenges the agent's ability to apply precise control for coordinating multiple linked joints.
    \item CarRacing: The easiest control task to learn from pixels - a top-down racing environment. The generated track is random in every episode.
    \item Lunar Lander: It is a classic rocket trajectory optimization problem. According to Pontryagin’s maximum principle, it is optimal to fire the engine at full throttle or turn off. This is why this environment has discrete actions: engine on or off.
     \item Bipedal Walker: a two-legged robot attempting to walk across varied terrain. The goal is for the agent to learn how to navigate efficiently and avoid falling.
\end{enumerate}
\clearpage
\subsubsection{MetaDrive Block Type Description}
\label{blocktypedescription}
\begin{table}[htbp]
\centering
\caption{Block Types Used in Experiments}
\label{tab:block_types}
\begin{tabular}{|c|l|}
\hline
\textbf{ID} & \textbf{Block Type} \\ \hline
S & Straight \\ \hline
C & Circular \\ \hline
r & InRamp \\ \hline
R & OutRamp \\ \hline
O & Roundabout \\ \hline
X & Intersection \\ \hline
y & Merge \\ \hline
Y & Split \\ \hline
T & T-Intersection \\ \hline
\end{tabular}
\end{table}

\begin{figure}[htbp]
    \centering
    \includegraphics[width=\textwidth]{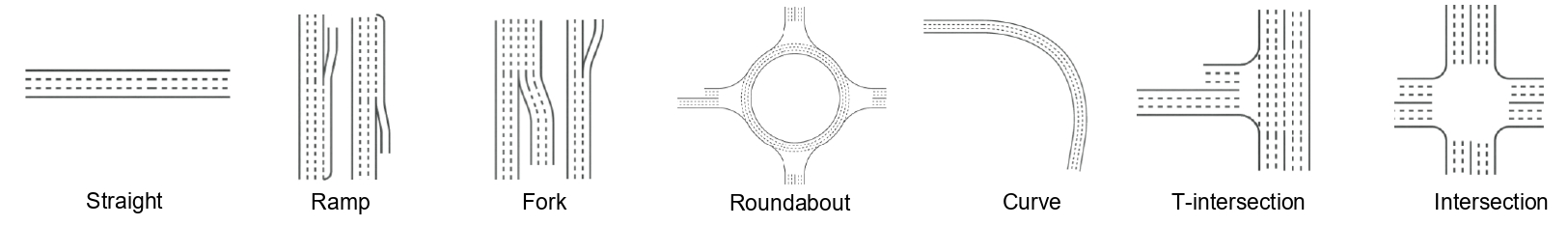}
    \caption{Various block types used in the MetaDrive environment. These blocks represent common road structures such as straight roads, ramps, forks, roundabouts, curves, T-intersections, and intersections, used for evaluating the vehicle's path planning and decision-making capabilities.}
    \label{fig:block_types}
\end{figure}

\subsubsection{Map Design and Testing Objectives}
\label{mapdesign}

\paragraph{Map 1: SrOYCTRyS}
This map consists of straight roads, roundabouts, intersections, T-intersections, splits, and ramps. The environment presents a highly complex combination of multiple intersections, dynamic traffic flow, and varying road structures.

\textbf{Testing Objective:} The focus of this environment is to evaluate the algorithm's smooth decision-making and multi-intersection handling, mimicking human driving behavior. The challenges include adjusting vehicle paths in real-time and ensuring smooth lane transitions in the presence of complex road structures such as roundabouts and ramps.

\paragraph{Map 2: COrXSrT}
This map combines circular roads, roundabouts, straight roads, intersections, ramps, and T-intersections. The environment is designed to assess the vehicle's decision-making capabilities when dealing with continuous changes in road grades and multiple intersection types.

\textbf{Testing Objective:} This environment tests the algorithm's ability to dynamically adjust to \textbf{grade changes} and \textbf{multi-intersection interactions}, replicating human-like behavior. The goal is to observe how well the algorithm adjusts vehicle speed and direction, ensuring stability in scenarios involving ramps and complex road networks.

\paragraph{Map 3: rXTSC}
This map consists of ramps, intersections, T-intersections, straight roads, and circular roads. The environment simulates multiple road interactions, testing the vehicle's path selection and stability, particularly at intersections and ramps.

\textbf{Testing Objective:} This environment evaluates the algorithm's performance in handling intersections and T-junctions with real-time path selection. The challenge is to ensure human-like adaptability when encountering multiple directional options, maintaining decision stability in dynamic traffic situations.

\paragraph{Map 4: YOrSX}
This map includes splits, roundabouts, straight roads, circular roads, and intersections. The environment is tailored to test the vehicle's ability to make path decisions in high-speed settings, particularly when merging traffic and navigating through complex junctions.

\textbf{Testing Objective:} The map focuses on testing the vehicle's ability to handle \textbf{high-speed lane merging} and \textbf{dynamic path planning}. The algorithm must mimic human drivers by making real-time adjustments in a high-speed environment, choosing optimal paths while maintaining speed control and safety through complex intersections and roundabouts.

\paragraph{Map 5: XTOC}
This map features circular roads, T-intersections, and straight roads, creating a unique combination of continuous curves and abrupt directional changes. The environment presents the challenge of maintaining speed while negotiating tight turns and quick transitions at T-intersections.

\textbf{Testing Objective:} The focus is on testing the vehicle's ability to handle \textbf{sharp directional changes} and maintain control during high-speed maneuvers. The algorithm needs to balance speed with precision, ensuring safe navigation through tight turns and abrupt intersections.

\paragraph{Map 6: XTSC} This map features a T-shaped intersection with traffic signals controlling vehicle flow from three directions. It tests advanced driving skills including traffic light compliance, turn management, and interaction with vehicles from cross directions.

\textbf{Testing Objective:} The main challenge is to evaluate the vehicle’s ability to maintain \textbf{lane stability} and make appropriate \textbf{speed adjustments} while navigating long straight roads and transitioning into a circular roundabout. The algorithm must ensure smooth control and decision-making, simulating human-like behavior in handling both high-speed straight roads and slower, more controlled turns in the roundabout.

\paragraph{Map 7: TOrXS}
This map consists of T-intersections, roundabouts, straight roads, and splits, forming a compact yet intricate structure. The layout challenges the algorithm to manage dynamic path selection and adapt to sudden directional changes within a moderately complex road network.

\textbf{Testing Objective:} The primary objective is to evaluate the algorithm's ability to manage split paths and handle sudden directional changes. The map focuses on the vehicle's adaptability in navigating roundabouts and maintaining stability while making real-time path decisions at T-intersections.

\paragraph{Map 8: CYrXT} This map integrates circular roads, Y-intersections, ramps, T-intersections, and straight roads, creating a dynamic and highly interconnected network. The layout introduces varying road geometries and frequent directional changes, requiring seamless decision-making and adaptability.

\textbf{Testing Objective:} The map is designed to test the algorithm’s ability to adapt to sudden directional shifts at Y-intersections and T-junctions, maintain stability on ramps, and execute precise maneuvers on circular roads. The emphasis is on smooth transitions between road types, effective navigation through interconnected pathways, and robust handling of diverse traffic scenarios.




\newpage
\subsubsection{MuJoCo Environments}
\begin{figure}[!htbp]
    \centering
    \includegraphics[width=0.2\textwidth]{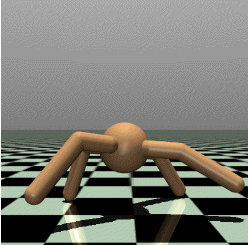}
    \includegraphics[width=0.2\textwidth]{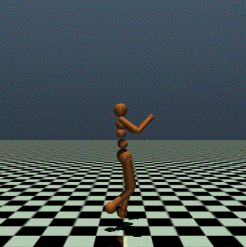}
    \includegraphics[width=0.2\textwidth]{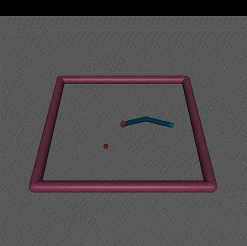}
    \includegraphics[width=0.2\textwidth]{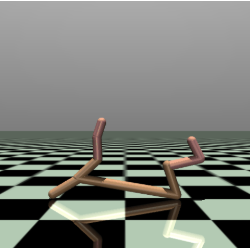}
    \caption{Ant, Humanoid, Reacher and Half Cheetah.}
    \label{fig:Continuous Env}
\end{figure}

\begin{enumerate}
    \item Ant: a 3D robot with a single central torso and four articulated legs is designed to navigate in the forward direction. The robot’s movement depends on coordinating the torque applied to the hinges that connect the legs to the torso and the segments within each leg.
    \item Humanoid: a 3D bipedal robot simulates human gait, with a torso, a pair of legs, and arms. Each leg and arm consists of two segments, representing the knees and elbows respectively; the legs are used for walking, while the arms assist with balance. The robot's goal is to walk forward as quickly as possible without falling.
    \item Humanoid Standup: The environment starts with the humanoid laying on the ground, and then the goal of the environment is to make the humanoid stand up and then keep it standing by applying torques to the various hinges.
    \item Reacher: a two-jointed robot arm. The goal is to move the robot’s end effector close to a target that is spawned at a random position.
    \item Half Cheetah: a 2-dimensional robot consisting of 9 body parts and 8 joints connecting them (including two paws). The goal is to apply torque to the joints to make the cheetah run forward (right) as fast as possible, with a positive reward based on the distance moved forward and a negative reward for moving backward. 
    \item Hopper: a two-dimensional one-legged figure consisting of four main body parts - the torso at the top, the thigh in the middle, the leg at the bottom, and a single foot on which the entire body rests. The goal is to make hops that move in the forward (right) direction by applying torque to the three hinges that connect the four body parts.
    \item Walker-2d: a two-dimensional bipedal robot consisting of seven main body parts - a single torso at the top (with the two legs splitting after the torso), two thighs in the middle below the torso, two legs below the thighs, and two feet attached to the legs on which the entire body rests. The goal is to walk in the forward (right) direction by applying torque to the six hinges connecting the seven body parts.
    \item Pusher: a multi-jointed robot arm that is very similar to a human arm. The goal is to move a target cylinder (called object) to a goal position using the robot’s end effector (called fingertip).
    \item Inverted Pendulum: The environment consists of a cart that can be moved linearly, with a pole attached to one end and having another end free. The cart can be pushed left or right, and the goal is to balance the pole on top of the cart by applying forces to the cart.
    \item Inverted Double Pendulum: The environment involves a cart that can be moved linearly, with one pole attached to it and a second pole attached to the other end of the first pole (leaving the second pole as the only one with a free end). The cart can be pushed left or right, and the goal is to balance the second pole on top of the first pole, which is in turn on top of the cart, by applying continuous forces to the cart.

\end{enumerate}

\newpage
\subsubsection{Atari Environments}
\begin{figure}[!htbp]
    \centering
    \includegraphics[width=0.2\textwidth]{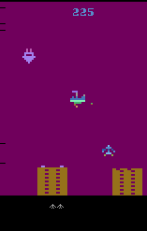}
    \includegraphics[width=0.2\textwidth]{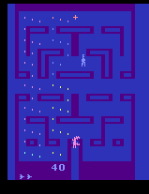}
    \includegraphics[width=0.2\textwidth]{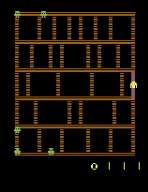}
    \includegraphics[width=0.2\textwidth]{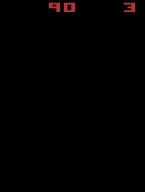}
    \includegraphics[width=0.2\textwidth]{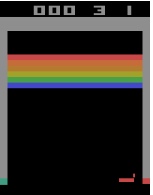}
    \includegraphics[width=0.2\textwidth]{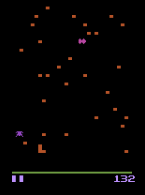}
    \includegraphics[width=0.2\textwidth]{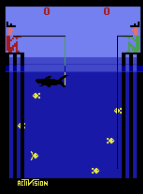}
    \includegraphics[width=0.2\textwidth]{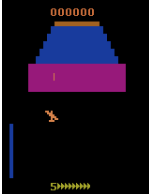}
    \caption{Air Raid, Alien, Amidar, Asteroids, Breakout, Centipede, Fishing Derby, Zaxxon.}
    \label{fig:Atari Env}
\end{figure}

\begin{enumerate}
    \item Air Raid: You control a ship that can move sideways and protect two buildings (one on the right and one on the left side of the screen) from flying saucers that are trying to drop bombs on them.
    \item Alien: You are stuck in a maze-like space ship with three aliens. You goal is to destroy their eggs that are scattered all over the ship while simultaneously avoiding the aliens (they are trying to kill you). 
    \item Admidar: You are trying to visit all places on a 2-dimensional grid while simultaneously avoiding your enemies. You can turn the tables at one point in the game: Your enemies turn into chickens and you can catch them.
    \item Asteroids: You control a spaceship in an asteroid field and must break up asteroids by shooting them. Once all asteroids are destroyed, you enter a new level and new asteroids will appear. You will occasionally be attacked by a flying saucer.
    \item Breakout: You move a paddle and hit the ball in a brick wall at the top of the screen. Your goal is to destroy the brick wall. You can try to break through the wall and let the ball wreak havoc on the other side, all on its own! You have five lives.
    \item Centipede: You are an elf and must use your magic wands to fend off spiders, fleas and centipedes. Your goal is to protect mushrooms in an enchanted forest.
    \item Fishing Derby: Your objective is to catch more sunfish than your opponent.
    \item Zaxxon: Your goal is to stop the evil robot Zaxxon and its armies from enslaving the galaxy by piloting your fighter and shooting enemies.
\end{enumerate}

\newpage

\end{document}